\documentclass[letterpaper,journal]{IEEEtran}
\usepackage{amsmath,amsfonts}
\usepackage{algorithmic}
\usepackage{algorithm}
\usepackage{array}
\usepackage[caption=false,font=normalsize,labelfont=sf,textfont=sf]{subfig}
\usepackage{textcomp}
\usepackage{stfloats}
\usepackage{url}
\usepackage{verbatim}
\usepackage{graphicx}
\usepackage{cite}
\usepackage{bm}
\usepackage{booktabs}
\usepackage{multirow}
\usepackage{colortbl}
\usepackage{makecell}
\definecolor{cvprblue}{rgb}{0.21,0.49,0.74}
\usepackage[normalem]{ulem}
\usepackage[T1]{fontenc}
	\usepackage[colorlinks,citecolor=black,linkcolor=black,urlcolor=black]{hyperref}
\usepackage{nicematrix}
\useunder{\uline}{\ul}{}
\usepackage{adjustbox} 
\usepackage{amssymb, bbding}
\definecolor{colorTab}{rgb}{0.95,0.90,0.92}

\begin{document}
	
	\title{Spatial-Frequency Learning for \\ Scanning Transmission Electron Microscopy \\ Image Enhancement}
	
	\author{Hesong Li, Ying Fu,~\IEEEmembership{Senior Member, IEEE}, Ziqi Wu, Ruiwen Shao, and Tao Zhang
		\IEEEcompsocitemizethanks{
			\IEEEcompsocthanksitem Hesong Li and Ying Fu are with the School of Computer Science and Technology, Beijing Institute of Technology, Beijing, China. Email: lihesong2@bit.edu.cn, fuying@bit.edu.cn. \protect\\
			\vspace{-8pt}
			\IEEEcompsocthanksitem Ziqi Wu and Ruiwen Shao are with the School of Medical Science and Engineering, Beijing Institute of Technology, Beijing, China. Email: wuziqi@bit.edu.cn, rwshao@bit.edu.cn. \protect\\
			\vspace{-8pt}
			\IEEEcompsocthanksitem Tao Zhang is with the School of Communication Engineering, Hangzhou Dianzi University, Hangzhou, China. Email: tzhang@hdu.edu.cn. \protect\\
			\vspace{-8pt}
		}
	}
	
	\markboth{IEEE Transactions on Pattern Analysis and Machine Intelligence,~Vol.~XX, No.~X, Month~2026}%
	{Li \MakeLowercase{\textit{et al.}}: Spatial-Frequency Learning for Scanning Transmission Electron Microscopy Image Enhancement}
	
	
	\maketitle
	
	\begin{abstract}
		Scanning Transmission Electron Microscopy (STEM) enables the observation of atomic arrangements at sub-angstrom resolution and plays a key role in the discovery and characterization of advanced materials. Raw STEM images are often noisy and blurred due to multiple degradations. Although recent deep learning methods improve STEM images, their performance remains limited under severe degradations. In this work, we propose a spatial-frequency learning framework for STEM image enhancement. Specifically, we design a spatial-frequency fusion network that exploits periodic atomic arrangements in the frequency domain and restores local atomic details in the spatial domain. Spatial- and frequency-domain features are enhanced separately and then fused. We also present a STEM noise calibration method for synthesizing more realistic STEM images. The parameters of scan noise, pointwise noise, and background noise are estimated through statistical analysis and fitting of real STEM images. Using these calibrated parameters, we construct a diverse dataset containing a wider range of atomic arrangements, imaging modes, and degradations. Random masks combine different atomic arrangements to simulate interfaces between multiple materials. Experimental results demonstrate that our method achieves state-of-the-art enhancement performance on synthetic datasets and real images of advanced materials while improving atomic localization accuracy.
	\end{abstract}
	
	
	\section{Introduction}
	\label{sec:intro}
	Scanning Transmission Electron Microscopy (STEM) enables scientific observation at the atomic scale and plays a crucial role in the discovery and study of materials. Recent studies have discovered new properties of advanced materials using STEM, including semiconductor materials \cite{semiconductor,semiconductor2}, superconducting materials \cite{superconduct}, new energy materials \cite{Battery1,Battery2,Battery3}, functional materials \cite{functional,functional2}, neuromorphic computing materials \cite{Interfaces}, and nanodevices \cite{heat}. 
	
	Although STEM provides atomic-scale resolution, its images are susceptible to multiple types of noise \cite{abTEM, AtomSegNet}. These include scan noise caused by probe jitter, pointwise noise introduced during signal readout, and background interference caused by impurities. These types of noise can overwhelm useful signals and hinder the observation of material structures. To avoid sample damage, both the exposure time and electron-beam intensity are constrained, further increasing the noise level. Moreover, spherical aberration blurs atomic contours and makes different atoms difficult to distinguish. Therefore, STEM image enhancement is needed to improve the visualization and analysis of material structures.
	
	\begin{figure}
		\centering
		\includegraphics[width=1\linewidth]{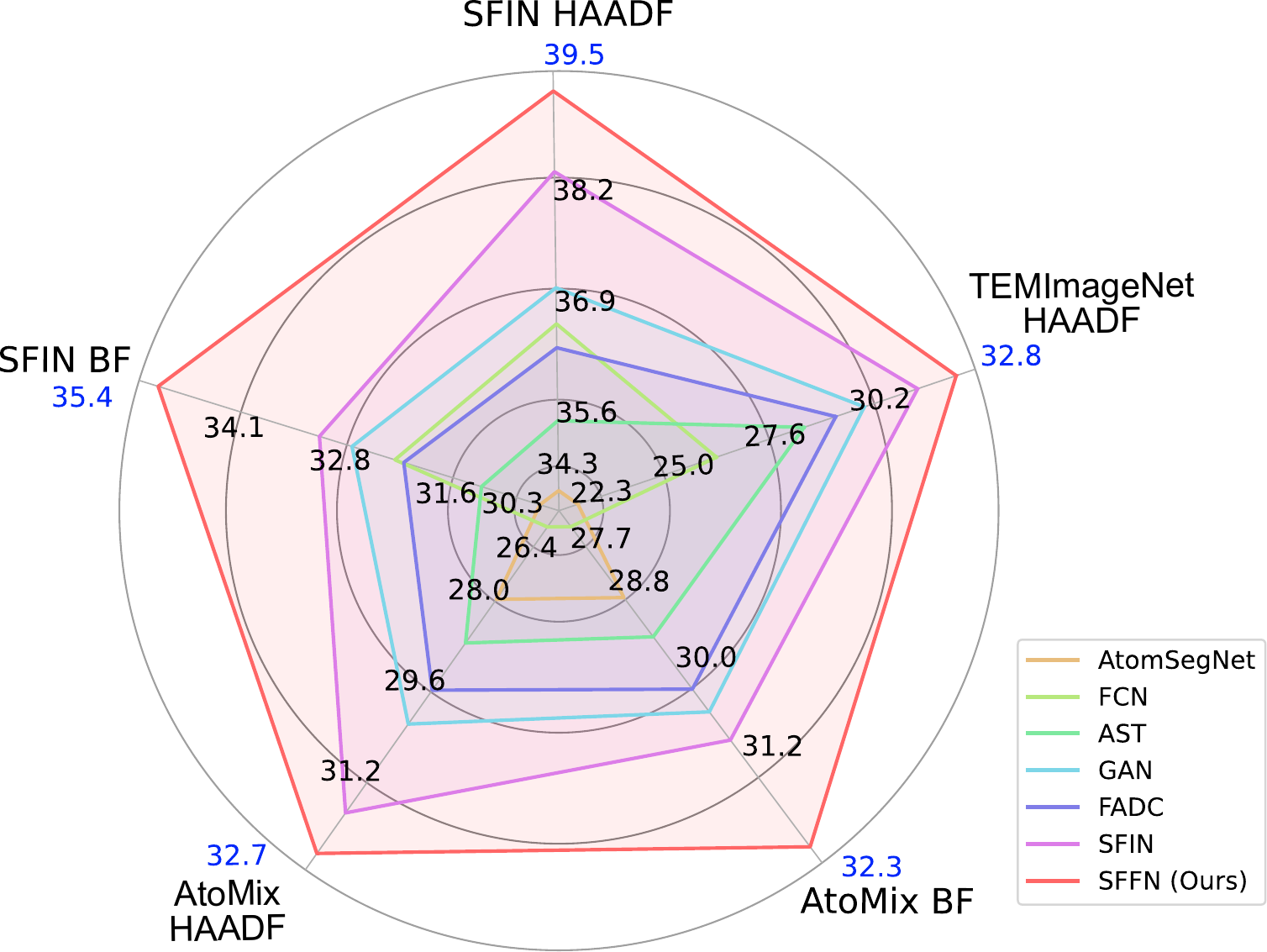}
		\vspace{-7mm}
		\caption{PSNR results of different methods on STEM image enhancement. Our spatial-frequency fusion network (SFFN) outperforms AtomSegNet \cite{AtomSegNet}, FCN \cite{FCN}, AST \cite{AST}, GAN \cite{GAN}, FADC \cite{FADC}, and SFIN \cite{SFIN} on three datasets (\emph{i.e.}, TEMImageNet \cite{AtomSegNet}, SFIN dataset \cite{SFIN}, and our AtoMix). HAADF and BF denote two imaging modes, namely high-angle annular dark-field imaging and bright-field imaging, respectively.}
		\vspace{-5mm}
		\label{fig:head}
	\end{figure}
	
	Simple image filtering methods (\emph{e.g.}, Gaussian filter, Wiener filter \cite{Wiener}, and bilateral filter \cite{Bilateral}) can partially remove noise, but their effectiveness is limited, and they cannot eliminate background interference and spherical aberration. Recently, some deep learning-based methods \cite{AtomSegNet,FCN,GAN, UDVD} have been designed for electron microscope image enhancement. Unsupervised methods (\emph{e.g.}, UDVD \cite{UDVD}) typically utilize continuous multi-frame STEM images as training data. Since adjacent frames are used as supervision signals, they can only remove noise that varies across frames, but struggle to eliminate background and spherical aberration, which persist in every frame. By synthesizing simulated noisy images and ground truth images, supervised methods (\emph{e.g.}, AtomSegNet \cite{AtomSegNet}, FCN \cite{FCN}, and GAN \cite{GAN}) can simultaneously remove noise, background, and spherical aberration, and generate output images suitable for other tasks such as atomic localization. Therefore, the performance of neural networks and the quality of synthetic data determine the effectiveness of the STEM image enhancement methods.
	
	One challenge is that STEM image signals are often overwhelmed by noise, making accurate feature extraction difficult. Existing networks \cite{AtomSegNet,FCN} mainly process spatial-domain features and neglect the strong prior induced by periodic atomic arrangements in the frequency domain. To address this issue, we design a spatial-frequency fusion network for STEM image enhancement. Our network restores atomic contours in the spatial domain while exploiting frequency-domain priors to suppress noise and enhance periodic structures. It processes features in both domains and adaptively fuses their complementary information. As shown in Fig.~\ref{fig:head}, our network outperforms existing state-of-the-art methods.
	
	The second challenge is the lack of realistic data for training deep learning models. On one hand, the prolonged exposure can damage the material sample and does not eliminate background and spherical aberration, so the ideal clean STEM images as supervised labels cannot be obtained through actual imaging. On the other hand, existing synthetic datasets \cite{abTEM,AtomSegNet,GAN-data} lack reality in noise distribution. STEM images have unique scan noise related to image gradients, making traditional calibration methods requiring uniform calibration frames \cite{calib1,calib2} unavailable. Thus, existing datasets usually use random noise parameters, which are different from those in real STEM images. In this work, we design a STEM image calibration method capable of getting various noise parameters from STEM images with atomic patterns, which does not require uniform calibration frames. We calibrate the noise parameters in real STEM images and apply them to synthesize noisy images, thereby ensuring the reality of the simulated STEM images in our dataset.
	
	Existing datasets also lack variety, and their scope of application is limited. First, they only include regular atomic patterns and exclude random arrangements, which often appear at material interfaces and are key parts in observing material changes. They also lack images with multiple arrangements. Second, they only include High-Angle Annular Dark-Field (HAADF) mode images, missing Bright-Field (BF) mode images, which are more suitable for light atoms observation \cite{bf1, bf2,bf3}. Third, most datasets lack simulations of more complex degradations, including spherical aberration, lower atomic brightness, and lower resolution. To address these issues, we construct a new dataset called AtoMix, which contains images with \textbf{mix}ed \textbf{ato}mic arrangements. It includes simulations of the various imaging modes, atom arrangements, and degradations mentioned earlier, ensuring data diversity. Our dataset includes complex degraded images that are closer to real-world data.

	Experimental results show that both our spatial-frequency fusion network and our AtoMix dataset are effective for STEM image enhancement. Our method makes STEM images clearer and atomic localization more accurate. The application results on various advanced materials demonstrate the effectiveness of our method and its role in material research.
	
	In summary, our main contributions are that we
	\begin{itemize}
		\item Present a spatial-frequency fusion network, which can flexibly combine spatial-domain and frequency-domain information to enhance the features of STEM images.
		\item Design a STEM noise calibration method, which can be used to synthesize more realistic STEM images.
		\item Build a more diverse dataset that contains more types of atom arrangements, imaging modes, and degradations.
	\end{itemize}

	This paper is an extension of a preliminary conference paper SFIN \cite{SFIN}. Table~\ref{tab:diff} summarizes the main extensions from the conference version to this journal version. In terms of network architecture, we design the Spatial-Frequency Fusion Network (SFFN), which replaces the fixed spatial-frequency addition in SFIN with learnable spatial-frequency fusion and further introduces frequency-position guidance for fine-grained frequency-domain processing. In terms of dataset, we build AtoMix by expanding the atom patterns from 15 patterns from 5 crystal systems to 46 patterns from 7 crystal systems, introducing mixed atomic arrangements for simulating material interfaces, and incorporating stronger degradations, including spherical aberration, lower brightness, and lower resolution. These extensions enable our method to achieve stronger enhancement performance on synthetic datasets and real data of advanced materials. In addition, we add comparisons with more methods, including transformers, unsupervised methods, and other frequency-domain processing networks, as well as more visual analyses of frequency-domain processing and a detailed introduction to our data synthesis method.
	
\begin{table}
	\small
	\centering
	\caption{Comparison of main extensions from the conference version to the journal version.}
	\vspace{-2mm}
	\resizebox{\columnwidth}{!}{%
		\setlength{\tabcolsep}{0.8mm}
		\renewcommand{\arraystretch}{1.2}
		\begin{NiceTabular}{
				>{\centering\arraybackslash\columncolor[HTML]{FFFFFF}}m{1.8cm}
				>{\centering\arraybackslash\columncolor[HTML]{FFFFFF}}m{2.7cm} |
				>{\centering\arraybackslash\columncolor[HTML]{FFFFFF}}m{2.8cm} |
				>{\centering\arraybackslash\columncolor[HTML]{FFFFFF}}m{3.2cm}
			}
			\hline
			\multicolumn{2}{c|}{\cellcolor[HTML]{FFFFFF}Category}
			& Conference version
			& Journal version \\ \hline
			
			\Block{2-1}{\hspace*{-0.85cm}\makebox[1.8cm][c]{Network}}
			& \parbox[c][1.00cm][c]{2.7cm}{
				\centering Spatial-frequency interaction
			}
			& \parbox[c][1.00cm][c]{2.8cm}{
				\centering Fixed addition
			}
			& \parbox[c][1.00cm][c]{3.2cm}{
				\centering
				Learnable spatial-frequency\\
				fusion
			} \\ \cline{2-4}
			
			& \parbox[c][0.90cm][c]{2.7cm}{
				\centering Frequency-position guidance
			}
			& \parbox[c][0.90cm][c]{2.8cm}{
				\centering Not included
			}
			& \parbox[c][0.90cm][c]{3.2cm}{
				\centering Included
			} \\ \hline
			
			\Block{3-1}{\hspace*{-0.85cm}\makebox[1.8cm][c]{Dataset}}
			& \parbox[c][0.90cm][c]{2.7cm}{
				\centering Atom patterns
			}
			& \parbox[c][0.90cm][c]{2.8cm}{
				\centering
				15 patterns from\\
				5 crystal systems
			}
			& \parbox[c][0.90cm][c]{3.2cm}{
				\centering
				46 patterns from\\
				7 crystal systems
			} \\ \cline{2-4}
			
			& \parbox[c][1.50cm][c]{2.7cm}{
				\centering Atom arrangements
			}
			& \parbox[c][1.50cm][c]{2.8cm}{
				\centering
				Regular, random\\
				addition, and\\
				random missing
			}
			& \parbox[c][1.50cm][c]{3.2cm}{
				\centering
				Regular, random\\
				addition, random\\
				missing, and mixed\\
				arrangements
			} \\ \cline{2-4}
			
			& \parbox[c][2.10cm][c]{2.7cm}{
				\centering Degradation
			}
			& \parbox[c][2.10cm][c]{2.8cm}{
				\centering
				Noise (background\\
				noise $\boldsymbol{N}_{\mathrm{background}}$,\\
				scan noise $\boldsymbol{N}_{\mathrm{scan}}$,\\
				and pointwise noise\\
				$\boldsymbol{N}_{\mathrm{pointwise}}$)
			}
			& \parbox[c][2.10cm][c]{3.2cm}{
				\centering
				Noise ($\boldsymbol{N}_{\mathrm{background}}$,\\
				$\boldsymbol{N}_{\mathrm{scan}}$, and\\
				$\boldsymbol{N}_{\mathrm{pointwise}}$),\\
				spherical aberration,\\
				lower brightness, and\\
				lower resolution
			} \\ \hline
		\end{NiceTabular}%
	}
	\label{tab:diff}
	\vspace{-2mm}
\end{table}
	
	\section{Related Work}
	\label{sec:rw}
	
	In this section, we provide an overview of the related work, including STEM image enhancement, frequency domain networks, and STEM image synthesis.
	
	\vspace{2pt}\noindent\textbf{STEM image enhancement.} Natural-image enhancement and recognition \cite{FCDFusion,CJE1,CJE2,CJE3,CJE4,CJE5, he1, he2, chen1, chen2, chen3,pl1,pl2,pl3,pl4,zyk2,lyp1,lyp2,TriMSOD} have been widely studied. Transformer-based methods \cite{xj1,xj2,JIG,zyk1,zyk4} and diffusion models \cite{he3,zyk3,lmy} leverage long-range dependencies and degradation modeling to achieve superior image restoration performance. For STEM images, three commonly used traditional enhancement methods are the Gaussian filter, Wiener filter \cite{Wiener}, and bilateral filter \cite{Bilateral}, which have been integrated into electron microscopy data processing software such as Gatan Microscopy Suite. These methods can suppress noise but cannot remove background interference or spherical aberration. Recently, deep learning-based methods \cite{AtomSegNet,FCN,GAN,SFIN,lhs-small} have achieved better enhancement performance. A fully convolutional network \cite{FCN} enhances images and identifies atomic defects, including vacancies and substitutions. A generative adversarial network \cite{GAN} is pretrained on the large-scale cell electron microscopy dataset CEM500k \cite{CEM500k} and fine-tuned for downstream tasks such as STEM image enhancement using the TEMImageNet dataset \cite{AtomSegNet}. AtomSegNet \cite{AtomSegNet} is a convolutional neural network based on the U-Net architecture \cite{UNet}; it is trained on TEMImageNet \cite{AtomSegNet} for STEM image enhancement and atomic localization. AtomSegNet has played a key role in studies of new battery materials \cite{Battery1,Battery2,Battery3}. Several unsupervised methods (\emph{e.g.}, UDVD \cite{UDVD}, 3D-B2U \cite{3D-B2U}, and ZeroIG \cite{ZeroIG}) have been proposed for denoising electron microscopy or natural images. Although these methods do not require synthetic training data, the absence of ideal ground-truth supervision limits their ability to remove background interference and correct atomic-contour distortions. Existing STEM image enhancement methods \cite{AtomSegNet,FCN,GAN} operate primarily in the spatial domain and therefore fail to exploit the strong frequency-domain information inherent in periodic atomic arrangements.

\vspace{2pt}\noindent\textbf{Frequency-domain networks.}
Image signals and noise typically lie in different frequency ranges. Recent studies in natural image processing show that incorporating frequency-domain processing into networks helps distinguish image signals from noise and recover image details. DeepRFT \cite{DeepRFT} employs Fourier transformation for image deblurring, FSNet \cite{FSNet} dynamically decomposes and selects frequency-domain features, and FADC \cite{FADC} and FDConv \cite{FDConv} use frequency analysis to guide convolutional feature processing. These existing frequency-domain methods mainly use frequency information to guide spatial-domain processing or perform frequency selection, rather than directly modeling individual frequency locations. In contrast, our SFFN directly processes features in the Fourier domain and introduces frequency-position guidance to distinguish different diffraction spots. Existing coordinate-guided methods, such as CoordConv \cite{CoordConv} and Coordinate Attention \cite{CoordAttention}, inject spatial coordinates into spatial-domain features, whereas our SFFN embeds frequency coordinates into Fourier-domain features to distinguish diffraction spots with different directions and periods. Existing methods either use fixed addition for spatial-frequency interaction (SFIN \cite{SFIN} and DeepRFT \cite{DeepRFT}) or lack explicit cross-domain interaction (FSNet \cite{FSNet}, FADC \cite{FADC}, and FDConv \cite{FDConv}), whereas our SFFN uses learnable spatial-frequency fusion to adaptively combine periodic frequency-domain structures with local spatial anomalies for fine-grained denoising of atomic-scale STEM images.

	\vspace{2pt}\noindent\textbf{STEM image synthesis.} Due to the difficulty of obtaining ideal supervised labels in actual experiments, STEM image synthesis provides data support for deep learning-based STEM image enhancement methods \cite{zyh,jyq}. The abTEM code library \cite{abTEM} provides a noise-free STEM image rendering method that supports multiple imaging modes, but is slow and requires atomic structure files. TEMImageNet dataset \cite{AtomSegNet} contains simulated STEM images in HAADF mode and label images for STEM image enhancement and atomic localization tasks. It includes several fixed shooting field angles and rotation angles and incorporates scan noise, pointwise noise, and background with random parameters. Cycle-GAN \cite{GAN-data} is used to learn the mapping relationship between noiseless simulated STEM images and noisy real STEM images and generate more realistic STEM noise. This method relies entirely on reference STEM images and lacks flexibility in controlling various parameters in the data synthesis program. Existing synthetic datasets \cite{AtomSegNet, GAN-data} lack realistic STEM image noise and diverse atomic arrangement patterns, imaging modes, and image degradation types. This results in poor performance of the trained networks when applied to complex real-world data.

	\section{Method}
	\label{sec:data}
	
	In this section, we first introduce the task formulation and our motivation. Then, we present our spatial-frequency fusion network. Finally, we introduce our STEM image data synthesis method, which includes clean image generation, noise calibration, and noise generation.
	
	\subsection{Formulation and Motivation}
	\label{sec:moti}
	A STEM image with noise can be represented as the sum of a distorted clean image and various types of noise, \emph{i.e.},
	\begin{equation}\label{eq:Inoisy}
		\boldsymbol{I}_\text{noisy}=\text{SA}(\boldsymbol{I}_\text{clean})+ \boldsymbol{N}_\text{background}+\boldsymbol{N}_\text{scan}+\boldsymbol{N}_\text{pointwise},
	\end{equation}
	where $\boldsymbol{I}_\text{noisy}$ and $\boldsymbol{I}_\text{clean}$ denote the noisy and ideal clean images, respectively; $\mathrm{SA}(\cdot)$ denotes the spherical-aberration degradation operator; and $\boldsymbol{N}_\text{background}$, $\boldsymbol{N}_\text{scan}$, and $\boldsymbol{N}_\text{pointwise}$ denote background, scan, and pointwise noise, respectively.
	
	The additive formulation in Eq.~\eqref{eq:Inoisy} is grounded in the physical mechanisms of STEM imaging. $\boldsymbol{N}_\text{background}$ arises from sample contamination, which is independent of the atomic signal and is therefore modeled as additive noise. For $\boldsymbol{N}_\text{scan}$, arising from probe jitter, and $\boldsymbol{N}_\text{pointwise}$, arising from the detector's readout electronics and dark current, we retain additive formulations for computational tractability but explicitly account for their signal dependence by modeling their variances as functions of the local image gradient and pixel intensity, respectively. These heteroscedastic additive formulations reflect the empirical noise behavior while enabling independent calibration and stable deep learning optimization.
	
	STEM image enhancement aims to derive $\boldsymbol{I}_\text{clean}$ from $\boldsymbol{I}_\text{noisy}$. For HAADF-mode images, the ideal image background is pure black, while the brightest atomic centers are pure white. In contrast, BF-mode images have a pure white background with the most significant atomic centers appearing as pure black. STEM image enhancement requires preserving the relative brightness and size of different atoms in an image, which enables the distinction of various elements from the image.

	\begin{figure}
		\centering
		\includegraphics[width=1\linewidth]{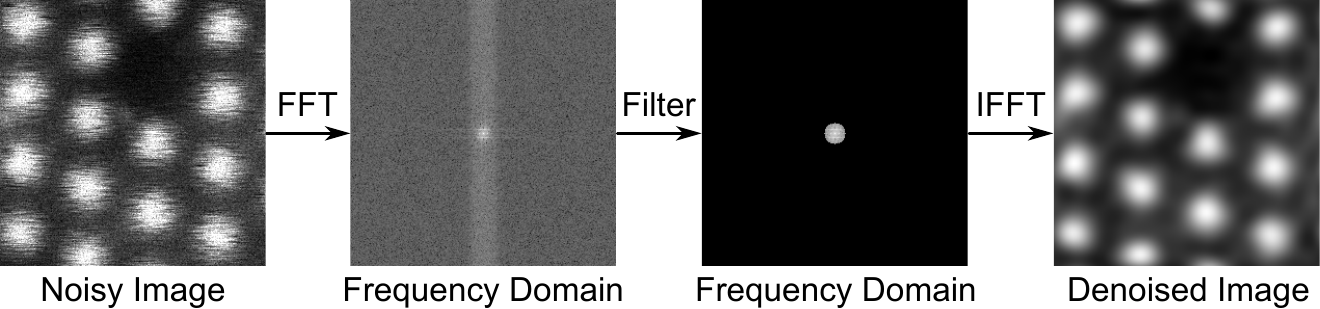}
		\vspace{-7mm}
		\caption{Motivation of our spatial-frequency fusion network. FFT and IFFT stand for Fast Fourier Transform and Inverse Fast Fourier Transform, respectively. A simple filter in the frequency domain can achieve good denoising effects for STEM images.}
		\vspace{-2mm}
		\label{fig:fft}
	\end{figure}
	
	Guided by the challenges outlined in the Introduction, our methodology is driven by two core motivations. First, while the periodic atomic lattice yields strong frequency signatures that can be leveraged for noise suppression, aperiodic features (\emph{e.g.}, defects, interfaces, or ion embeddings) remain spatially localized and cannot be fully characterized in the frequency domain alone. This necessitates a spatial-frequency fusion strategy that jointly enhances complementary information from both domains. To leverage the distinctive spot patterns of STEM images in the frequency domain, we embed coordinate information into the frequency-domain processing to enable fine-grained handling of information across different frequency bands. Second, robust model generalization requires training data that faithfully replicates real-world degradation characteristics. Since STEM scanning noise is intrinsically gradient-dependent, conventional flat-field calibration is inapplicable, motivating our novel parameter extraction scheme. Furthermore, to bridge the simulation-to-experiment gap, our synthesized dataset explicitly incorporates mixed atomic arrangements, dual imaging modes (HAADF/BF), and complex degradations (\emph{e.g.}, spherical aberration and intensity loss). The detailed network architecture and data synthesis pipeline are elaborated in Sections~\ref{sec:network-design} and \ref{sec:data-gen}, respectively.

	\subsection{Spatial-Frequency Fusion Network}
	\label{sec:network-design}
	As discussed in Section~\ref{sec:moti}, jointly modeling spatial- and frequency-domain information enables the network to capture both periodic atomic arrangements and local irregularities, thereby improving image enhancement. We next describe the overall network pipeline, the spatial-frequency fusion block, and the coordinate-guided frequency convolution used in each block.
	
\begin{figure}
		\centering
		\includegraphics[width=\linewidth]{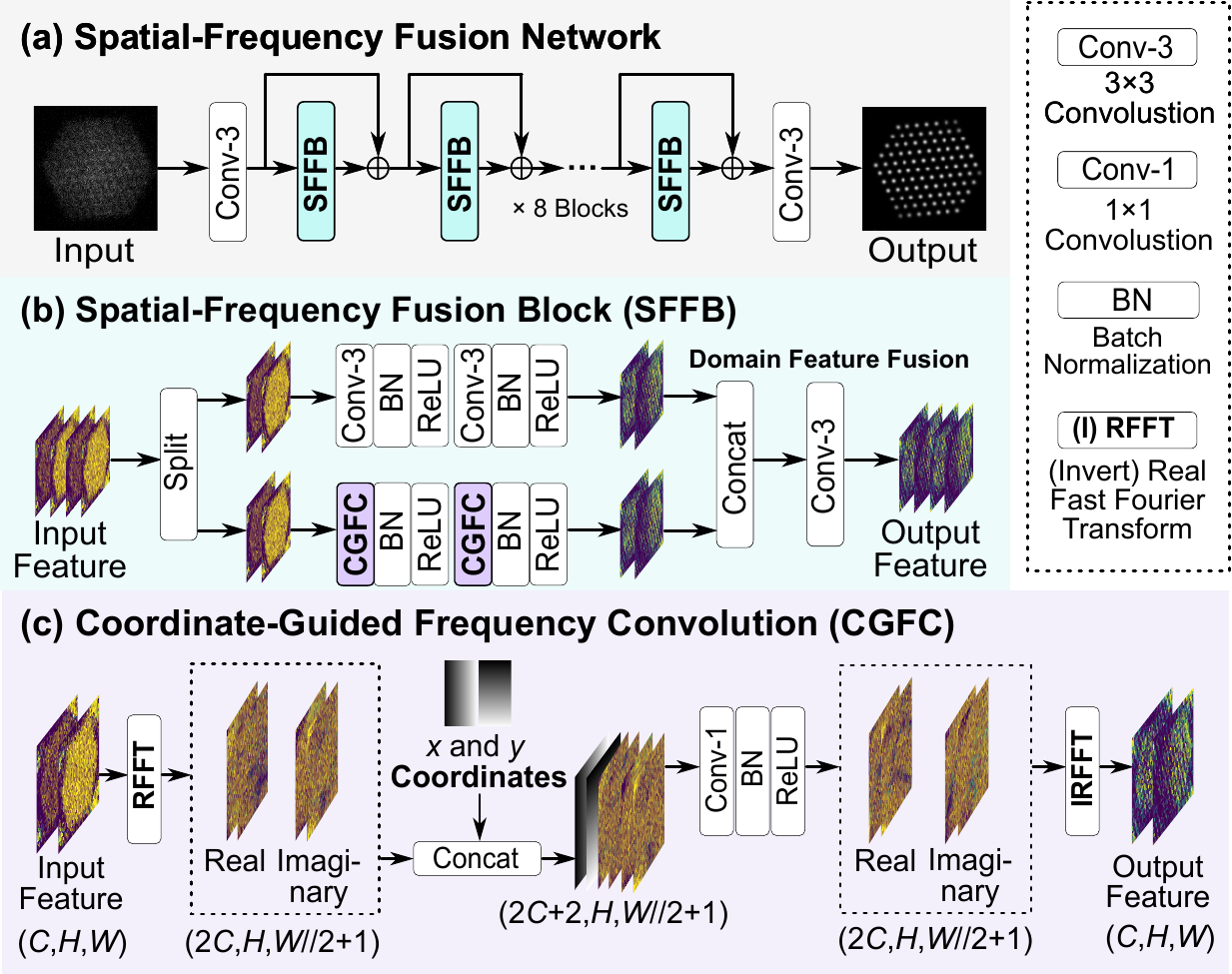}
		\vspace{-7mm}
		\caption{Framework of our Spatial-Frequency Fusion Network (SFFN). In each Spatial-Frequency Fusion Block (SFFB), the input features are processed by spatial- and frequency-domain branches, whose outputs are fused through a convolution. The Coordinate-Guided Frequency Convolution (CGFC) transforms features into the frequency domain using a real-valued Fast Fourier Transform (RFFT) and processes different frequency components with two coordinate channels. ``//'' denotes floor division.}
		\label{fig:frame}
		\vspace{-1mm}
	\end{figure}

	\vspace{2pt}\noindent\textbf{Overall pipeline.}
	Our network follows the basic architecture of common image restoration networks \cite{Restormer,SwinIR,AST}, including two convolutions at the head and tail, and multiple blocks stacked in the middle, as shown in Fig.~\ref{fig:frame}(a). First, a 3$\times$3 convolution extracts shallow features from the single-channel input image to form a feature map with $C$ channels. Multiple subsequent stacked spatial-frequency fusion blocks gradually enhance these features, where each block uses residual connections to stabilize training. Finally, a 3×3 convolution fuses the enhanced features into a single-channel output image. 
	
	\vspace{2pt}\noindent\textbf{Spatial-frequency fusion block.}
	As shown in Fig.~\ref{fig:frame}(b), we design the spatial-frequency fusion block as the basic module of our network, which can simultaneously enhance the spatial and frequency domain information in the feature maps. First, the feature map with $C$ channels is evenly divided into two parts along the channel dimension, each with $C/2$ channels. These two parts of features are then directed to two separate branches for enhancement in the spatial and frequency domains, respectively. The spatial domain branch employs a common convolutional module for enhancement, consisting of two 3$\times$3 convolutions, two batch normalizations, and two ReLU (Rectified Linear Unit) activation functions. In the frequency domain branch, the 3×3 convolution is replaced by a specially designed coordinate-guided frequency convolution (Fig.~\ref{fig:frame}(c)) that transforms features to the frequency domain for enhancement. It will be introduced later.
	
	The spatial- and frequency-domain branches process complementary information. Periodic atomic arrangements produce strong frequency-domain signatures that can be enhanced by the frequency-domain branch. Moreover, because signal and noise occupy different frequency ranges, this branch can enhance atomic signals while suppressing noise. The spatial-domain branch processes localized aperiodic structures that are difficult to characterize solely in the frequency domain, including material interfaces, atomic vacancies, and intercalated ions. Fusing the outputs of the two branches enables more comprehensive feature enhancement.
	
	Unlike the fixed cross-domain addition used in SFIN \cite{SFIN}, we adopt a convolutional operation to fuse features from the two domains. Specifically, the outputs of the spatial- and frequency-domain branches are concatenated along the channel dimension and then fused by a 3$\times$3 convolution. Because the complementary information from the two domains does not necessarily follow an additive relationship, convolutional fusion provides greater flexibility. This change does not increase computational complexity or the parameter count relative to SFIN \cite{SFIN}.
	
	\begin{figure*}
		\centering
		\includegraphics[width=.95\linewidth]{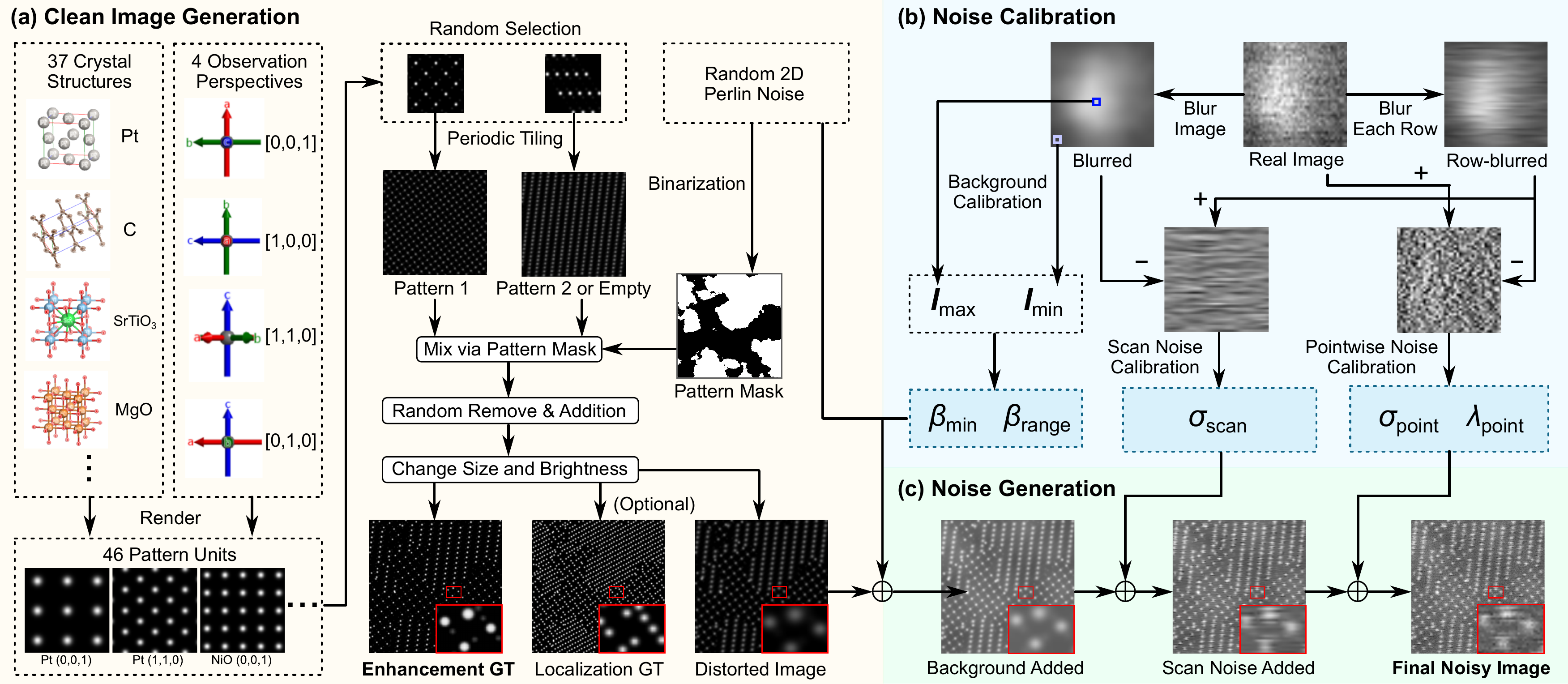}
		\vspace{-3mm}
		\caption{Our STEM image synthesis framework, including (a) clean image generation, (b) noise calibration, and (c) noise generation. Clean image generation obtains ground-truth labels and clean images without noise by mixing different atomic arrangement patterns via masks. Noise generation progressively adds various noises to the clean images, where the noise parameters are derived from the noise calibration of real images. Image noise parameters include: $\beta_\text{min}$ and $\beta_\text{range}$ are the minimum background brightness and brightness range, respectively; $\sigma_\text{scan}$ is the intensity of scan noise; $\sigma_\text{point}$ and $\lambda_\text{point}$ are the intensity and the distribution shape parameter of pointwise noise, respectively.}
		\label{fig:syn_frame}
		\vspace{-1mm}
	\end{figure*}
	
	\vspace{2pt}\noindent\textbf{Coordinate-guided frequency convolution.}
	We design the coordinate-guided frequency convolution to enhance features in the frequency domain. It distinguishes signal from noise by processing frequency-domain features augmented with frequency-position coordinates.
	
	First, the real-valued fast Fourier transform (\emph{i.e.}, the ``rfftn'' function in PyTorch \cite{PyTorch}) is used to convert each channel of the input feature map into the frequency domain separately. Since the input is a real-valued matrix, the transformed matrix exhibits conjugate symmetry, so only half of the information in the resulting feature map needs to be retained. As shown in Fig.~\ref{fig:frame}(c), the width of the resulting frequency-domain feature map is approximately halved. Due to the complex-valued result, the number of channels doubles, including both real and imaginary parts.
	
	To enable the convolution to distinguish different frequency locations, we append two coordinate maps to the frequency-domain features, representing the $x$ and $y$ coordinates of each location. A 1$\times$1 convolution then processes the frequency components while considering their positions, with inputs comprising the two coordinate channels and the remaining $2C$ channels of complex-valued features. Following a standard convolutional-module design, batch normalization and a ReLU activation are applied after the 1$\times$1 convolution. Finally, the enhanced frequency-domain features are transformed back to the spatial domain using the inverse real-valued fast Fourier transform (\emph{i.e.}, the ``irfftn'' function in PyTorch \cite{PyTorch}).
	
	\subsection{STEM Image Synthesis}
	\label{sec:data-gen}
	It is difficult to obtain ideal supervising labels for image enhancement from STEM imaging. To train the network using paired data, it is necessary to first synthesize ideal clean images and then add various types of degradations to simulate real noisy STEM images. As shown in Fig.~\ref{fig:syn_frame}, our STEM image synthesis framework includes clean image generation, noise calibration, and noise generation. Clean image generation obtains ground-truth labels and clean images without noise by mixing different atomic arrangement patterns via masks. Noise generation progressively adds various noises to the clean images, where the noise parameters are derived from the noise calibration of real images. The following are the details of each step in the order of the data synthesis process.

	\vspace{2pt}\noindent\textbf{Atom pattern generation from units.} 
	Materials observed by STEM are mostly crystalline structures, whose atomic arrangements exhibit significant periodicity. Therefore, the entire image can be formed by tiling repeated pattern units.
	
	First, we generate 46 pattern units, each representing a periodic unit of a complete image, as shown on the left side of Fig.~\ref{fig:syn_frame}(a). These pattern units are rendered from 37 crystal structures viewed from four orientations. Because of crystal symmetry, some structures produce the same pattern when viewed from different orientations. Our dataset covers the seven major crystal systems defined by the International Union of Crystallography (IUCr), more than previous datasets \cite{AtomSegNet, GAN-data}, thereby supporting a wider range of common material structures.
	
	Theoretically, a clean STEM image consists of numerous circular spots, each corresponding to an atomic column or a single atom and modeled using a 2D Gaussian distribution. The atomic species and column thickness determine the size and brightness of these spots. In HAADF images, atomic columns appear bright because the detector collects high-angle scattered electrons. In BF images, atomic columns appear dark because scattering reduces the intensity collected by the bright-field detector \cite{bf1,bf2,bf3}.
	The relative brightness and size of different atoms are fixed and can be simulated by the abTEM library \cite{abTEM}. Note that the abTEM library cannot be directly used to synthesize the STEM image dataset, as its physics-based rendering is slow and fails to flexibly synthesize images with different pattern combinations and various noises. In our data synthesis, abTEM is only used to obtain the arrangement structures of different pattern units, as well as the relative brightness and size parameters of atoms. These parameters are then used to render each atom with a 2D normal distribution.
	
	After randomly selecting a pattern unit, we periodically tile it to form an image with a consistent atomic arrangement. To diversify the orientations and scales of the synthesized data, the pattern unit is randomly rotated and scaled for each image. The resulting interatomic spacing ranges from approximately 7 to 30 pixels, based on measurements of real STEM images. The pattern unit is then tiled along two lattice directions, $\alpha_1$ and $\alpha_2$, to fill the image. Different pattern units have different $\alpha_1$ and $\alpha_2$, which are not necessarily orthogonal. By expressing the four image corners in the coordinate system defined by $\alpha_1$ and $\alpha_2$, we determine the number of repetitions required along each direction to cover the entire image.

	\begin{figure}
		\centering
		\includegraphics[width=.95\linewidth]{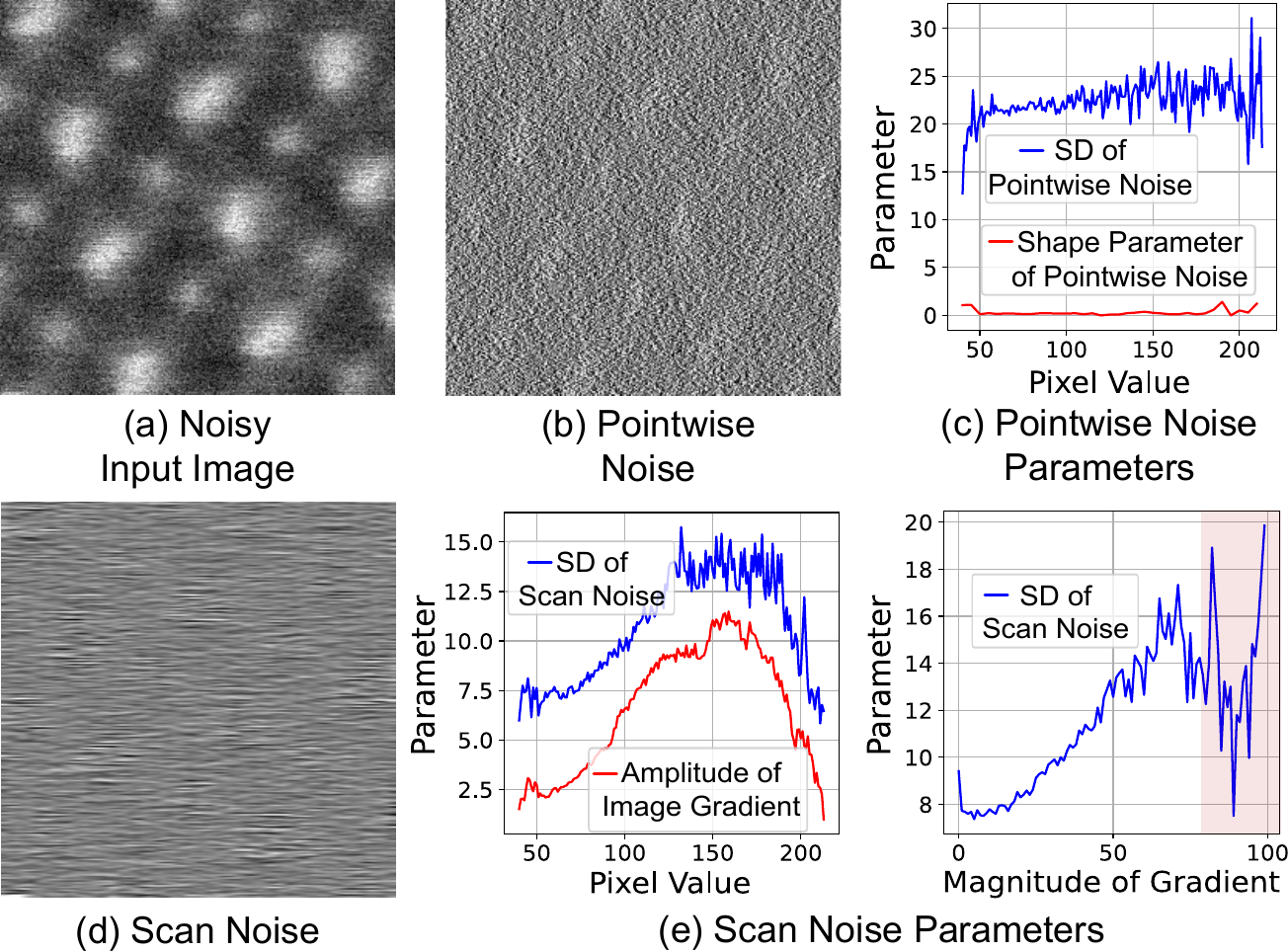}
		\vspace{-3mm}
		\caption{Pointwise noise (b) and scan noise (d) extracted from the noisy real input (a), together with their parameter variations in (c) and (e), respectively. ``SD'' denotes standard deviation. The light-red rectangular region in (e) indicates fluctuations caused by limited data and is excluded during fitting.}
		\label{fig:syn_calib}
		\vspace{-2mm}
	\end{figure}
	
	\vspace{2pt}\noindent\textbf{Mask-based pattern mixing.} Observing the interface between two atomic arrangements is important for analyzing interactions between materials and material growth. Therefore, the synthetic data should include combinations of multiple atomic arrangements with relatively random structures at their interfaces. Unlike previous datasets that simulate only one atomic arrangement per image, we use binarized Perlin noise \cite{Perlin} as a mask to mix two different arrangements. We first generate an image of one arrangement and replace the region where the mask equals 1 with another arrangement. The second arrangement is set to an empty arrangement with a probability of 0.1 to simulate the vacuum outside a material boundary.
	
	We then generate ground-truth and degraded images from the mixed atomic arrangements. To facilitate the adjustment of atomic brightness and size, each arrangement is represented as a list of per-atom attributes, including coordinates, brightness, and size, rather than being directly rendered as an image. To simulate atomic defects and ion intercalation, atoms are randomly removed from and added to the mixed arrangement. We approximate this process by relocating about 10\% of the atoms to random positions. When generating ground-truth images for enhancement, the atomic sizes remain unchanged, while the brightness values are proportionally scaled so that the brightest atom has an intensity of 255 in HAADF mode. When generating ground truth for atomic localization, each atom is rendered as a disk with a radius of 1 pixel and an intensity of 255.
	
	When generating a degraded image, the radius of each atom is doubled to approximate spherical-aberration-induced probe broadening. Spherical aberration in STEM originates from lens imperfections that broaden the probe and effectively widen its point spread function. Because the point spread function is commonly approximated by a 2D Gaussian distribution, increasing the Gaussian radius mimics this probe-broadening effect, which appears as blurred and enlarged atomic columns in the spatial domain. In addition, the intensity of the entire image is multiplied by a random value between 0.2 and 1 to simulate signal attenuation.
	
	Next, background, scan noise, and pointwise noise are sequentially added to the degraded image to form the final noisy image (Fig.~\ref{fig:syn_frame}(c)). The noise parameters are obtained by calibrating real images (Fig.~\ref{fig:syn_frame}(b)).
	
	\vspace{2pt}\noindent\textbf{Background calibration and generation.} 
	Background $\boldsymbol{N}_\text{background}$ comes from non-uniform sample thickness or impurities such as carbon compounds after long-term electron beam exposure. It can be simulated using 2D Perlin noise \cite{Perlin}. Since the background varies smoothly at different spatial positions, for a small region in a noiseless HAADF image, the minimum pixel value of the region can represent the local background brightness, while the maximum value represents the atomic brightness. Therefore, by statistically analyzing the minimum values of different regions in real images after Gaussian blurring, the variation range of background brightness in the HAADF mode can be obtained. In contrast, for the BF mode, the background brightness is the maximum value of the pixels in the region. 
	
	As shown in Fig.~\ref{fig:syn_frame}(b), small-sized patches (\emph{e.g.}, 32$\times$32)  $\boldsymbol{I}'$ are cropped from the blurred image to calculate the maximum and minimum values $\boldsymbol{I}_\text{max}$ and $\boldsymbol{I}_\text{min}$ for each location. Then, the background brightness parameters $\beta_\text{min}$ and $\beta_\text{range}$ can be calculated as
	\begin{equation}\label{eq:bmin}
		\beta_\text{min} = 
		\begin{cases} 
			\min\limits_{\boldsymbol{I}'} \boldsymbol{I}_\text{min} & \text{HAADF mode}, \\
			255 - \max\limits_{\boldsymbol{I}'} \boldsymbol{I}_\text{max} & \text{BF mode},
		\end{cases} 
	\end{equation}
	\begin{equation}
		\beta_\text{max} = 
		\begin{cases} 
			\max\limits_{\boldsymbol{I}'} \boldsymbol{I}_\text{min} & \text{HAADF mode}, \\
			255 - \min\limits_{\boldsymbol{I}'} \boldsymbol{I}_\text{max} & \text{BF mode},\label{eq:bmax}
		\end{cases} 
	\end{equation}
	\begin{equation}
		\beta_\text{range} =  \beta_\text{max} - \beta_\text{min}, \label{eq:brange}
	\end{equation}
	where $\boldsymbol{I}'$ represents a small block in the blurred image. The final synthetic background is
	\begin{equation}
		\boldsymbol{N}_\text{background} = 
		\begin{cases} 
			\beta_\text{min}\boldsymbol{1} + r\beta_\text{range}\boldsymbol{N}_\text{p} & \text{HAADF mode}, \\
			-\beta_\text{min}\boldsymbol{1} - r\beta_\text{range}\boldsymbol{N}_\text{p} & \text{BF mode},
		\end{cases}\label{eq:Nback}
	\end{equation}
	where $\boldsymbol{1}$ is an all-ones matrix of the same size as the image and $\boldsymbol{N_\text{p}}$ is a 2D Perlin noise \cite{Perlin} between 0 and 1. $r$ is a random number generated for each image between 0 and 1, which controls the intensity of the background.

	\vspace{2pt}\noindent\textbf{Scan noise calibration and generation.} 
	Scan noise $\boldsymbol{N}_\text{scan}$ is primarily due to the jitter of the scanning position. This results in the obtained pixel capturing random areas near the ideal imaging location. When there is significant contrast variation in these areas (such as at the edge of an atom), the pixel value differences caused by the position shift become more pronounced. Therefore, the amplitude of scan noise is approximately proportional to the amplitude of the image gradient at that location. To calibrate its standard deviation $\sigma_\text{scan}$, it is first necessary to isolate the scan noise from the noisy STEM image. As illustrated in Fig.~\ref{fig:syn_frame}(b), each row of the original image is blurred using a one-dimensional Gaussian filter, resulting in a row-wise blurred image. In it, the pointwise noise is approximately removed by the row-wise blurring, leaving only background and scan noise. Then, the difference between the row-wise blurred image and the blurred image represents the scan noise image, as shown in Fig.~\ref{fig:syn_calib}(d). Fig.~\ref{fig:syn_calib}(e) presents the standard deviation of scan noise as a function of pixel brightness and image gradient amplitude, respectively. It can be observed that the standard deviation of scan noise is proportional to the amplitude of the image gradient. Thus, a linear fit can be applied to determine the slope and intercept of $\sigma_\text{scan}$ related to image gradient amplitude, which can then be used to simulate the scan noise.

\vspace{2pt}\noindent\textbf{Pointwise noise calibration and generation.} 
Pointwise noise $\boldsymbol{N}_\text{pointwise}$ refers to noise that is independent between individual pixels, originating from the detector's signal processing system \cite{calib1,calib2}, such as dark current and readout noise. Related studies on the calibration of noise in natural images \cite{calib1,calib2} have confirmed that Tukey's lambda distribution with a shape parameter provides a more accurate estimation of pointwise noise, \emph{i.e.},
\begin{equation}
	\boldsymbol{N}_\text{pointwise} \sim \text{TL}(0,\sigma_\text{point}, \lambda_\text{point}),
\end{equation}
where $\sigma_\text{point}$ is the standard deviation and $\lambda_\text{point}$ is the shape parameter. Pointwise noise can be obtained by calculating the difference between the original image and the row-wise blurred image, as shown in Fig.~\ref{fig:syn_calib}(b). Fig.~\ref{fig:syn_calib}(c) illustrates the variation of two parameters with pixel values. The standard deviation is proportional to the pixel value, while the shape parameter is a small positive constant number. Therefore, the slope and intercept of the standard deviation $\sigma_\text{point}$, along with the constant shape parameter $\lambda_\text{point}$, can be calculated from real images and used for the simulation of pointwise noise.

Our calibration framework is mode-agnostic and applicable to common STEM modes (\emph{e.g.}, HAADF, BF). While scan noise, pointwise noise, and background share consistent physical origins across modes, the contrast inversion between HAADF (bright atoms) and BF (dark atoms) requires simple sign adjustments in parameter extraction (Eqs.~\eqref{eq:bmin}\eqref{eq:bmax}\eqref{eq:brange}\eqref{eq:Nback}). For other imaging modes, the same calibration pipeline can be directly applied by fitting mode-specific parameters from representative real images. The SNR of our synthesized AtoMix dataset ranges from $-$5.09 dB to 10.58 dB across BF and HAADF modes, where negative values reflect the extremely low-dose conditions targeted by our work (\emph{i.e.}, noise power exceeding signal power in beam-sensitive imaging), ensuring the practical relevance of our dataset for training enhancement models under real-world low-SNR STEM scenarios.

\section{Experiments}
\label{sec:exp}

In this section, we first introduce the experimental setup. Then, we compare the enhancement results of different methods. Finally, we analyze in detail the effectiveness of the proposed spatial-frequency learning and our AtoMix dataset.

\subsection{Experimental Settings}
We first provide the datasets and metrics. Then, we introduce the comparison methods. Finally, we provide implementation details of our spatial-frequency fusion network. 

\vspace{5pt}\noindent\textbf{Datasets and metrics.} 
We validate the effectiveness of our method on both synthetic datasets and real-world data. The synthetic datasets include TEMImageNet \cite{AtomSegNet}, SFIN dataset \cite{SFIN}, and our AtoMix. The real-world data comprises publicly available images of four types of advanced materials (\emph{i.e.},  superconducting material \cite{superconduct}, all-solid-state battery \cite{Battery1}, nanodevice \cite{heat}, and neuromorphic computing material \cite{Interfaces}), as well as STEM images of high-entropy alloys captured in our laboratory. 
Following SFIN \cite{SFIN}, we use Peak Signal-to-Noise Ratio (PSNR) and Structural Similarity Index Measure (SSIM) \cite{SSIM} to evaluate enhancement performance.

\begin{figure*}
	\centering
	\includegraphics[width=.85\linewidth]{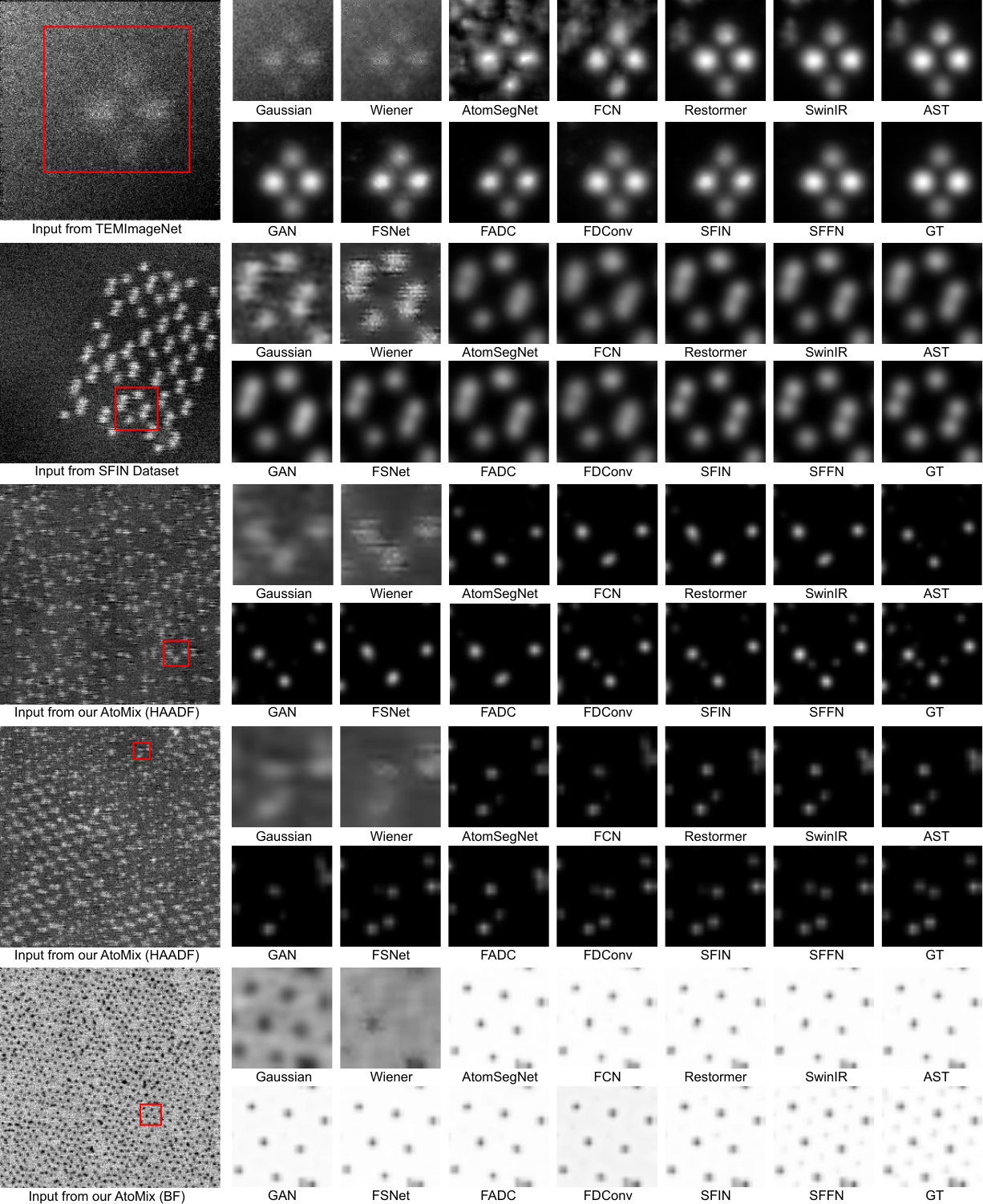}
	\vspace{-3mm}
	\centering
	\caption{Enhancement results on different synthetic datasets. The training data used corresponds to the training set associated with the test data.}
	\label{fig:main}
	\vspace{0mm}
\end{figure*}

\begin{table*}\small
	\centering
	\caption{Enhancement results on different synthetic datasets. Deep learning methods are trained and tested on the corresponding datasets. HAADF and BF are imaging modes. Best and second best values are \textbf{bold} and {\ul underlined}, respectively.}
	\vspace{-2mm}
	\resizebox{.95\textwidth}{!}{
		\setlength{\tabcolsep}{1.mm}
		\begin{NiceTabular}{c|c|cc|cccc|cccc}
			\hline
			&                                                                                           & \multicolumn{2}{c|}{TEMImageNet \cite{AtomSegNet}}                               & \multicolumn{4}{c|}{SFIN Dataset \cite{SFIN}}                                                                                                                                            & \multicolumn{4}{c}{AtoMix (Ours)}                                                                                                                                                        \\ \cline{3-12} 
			&                                                                                           & \multicolumn{2}{c|}{HAADF Mode}                                                  & \multicolumn{2}{c|}{HAADF Mode}                                                                       & \multicolumn{2}{c|}{BF Mode}                                                     & \multicolumn{2}{c|}{HAADF Mode}                                                                       & \multicolumn{2}{c}{BF Mode}                                                      \\
			\multirow{-3}{*}{Method}            & \multirow{-3}{*}{Category}                                                                & PSNR $\uparrow$                        & SSIM $\uparrow$                         & PSNR $\uparrow$                        & \multicolumn{1}{c|}{SSIM $\uparrow$}                         & PSNR $\uparrow$                        & SSIM $\uparrow$                         & PSNR $\uparrow$                        & \multicolumn{1}{c|}{SSIM $\uparrow$}                         & PSNR $\uparrow$                        & SSIM $\uparrow$                         \\ \hline
			Gaussian Filter \cite{Bilateral}    &                                                                                           & 16.78                                  & 0.4961                                  & 16.33                                  & \multicolumn{1}{c|}{0.3303}                                  & 10.30                                  & 0.6867                                  & 10.57                                  & \multicolumn{1}{c|}{0.1053}                                  & 10.17                                  & 0.5291                                  \\
			Wiener Filter \cite{Wiener}         & \multirow{-2}{*}{Traditional}                                                             & 16.08                                  & 0.4991                                  & 16.43                                  & \multicolumn{1}{c|}{0.3267}                                  & 10.23                                  & 0.8081                                  & 10.59                                  & \multicolumn{1}{c|}{0.0803}                                  & 10.01                                  & 0.4405                                  \\ \hline
			AtomSegNet \cite{AtomSegNet}        &                                                                                           & 22.34                                  & 0.8008                                  & 34.33                                  & \multicolumn{1}{c|}{0.8994}                                  & 30.26                                  & 0.9673                                  & 27.79                                  & \multicolumn{1}{c|}{0.8240}                                  & 28.70                                  & 0.8934                                  \\
			FCN \cite{FCN}                      &                                                                                           & 26.18                                  & 0.8842                                  & 32.82                                  & \multicolumn{1}{c|}{0.8830}                                  & 28.77                                  & 0.9623                                  & 26.38                                  & \multicolumn{1}{c|}{0.7711}                                  & 27.67                                  & 0.8765                                  \\
			Restormer \cite{Restormer}          &                                                                                           & 28.14                                  & 0.9113                                  & 34.82                                  & \multicolumn{1}{c|}{0.9099}                                  & 30.76                                  & 0.9745                                  & 28.28                                  & \multicolumn{1}{c|}{0.8373}                                  & 29.06                                  & 0.8945                                  \\
			SwinIR \cite{SwinIR}                &                                                                                           & 28.33                                  & 0.9066                                  & 35.00                                  & \multicolumn{1}{c|}{0.9085}                                  & 30.81                                  & 0.9696                                  & 28.39                                  & \multicolumn{1}{c|}{0.8387}                                  & 29.18                                  & 0.8984                                  \\
			AST \cite{AST}                      &                                                                                           & 28.61                                  & 0.9071                                  & 35.24                                  & \multicolumn{1}{c|}{0.9076}                                  & 31.05                                  & 0.9721                                  & 28.64                                  & \multicolumn{1}{c|}{0.8431}                                  & 29.27                                  & 0.9025                                  \\
			GAN \cite{GAN}                      & \multirow{-6}{*}{\begin{tabular}[c]{@{}c@{}}Spatial\\ Network\end{tabular}}               & 30.28                                  & 0.9311                                  & 36.97                                  & \multicolumn{1}{c|}{0.9347}                                  & 32.80                                  & 0.9759                                  & 30.22                                  & \multicolumn{1}{c|}{0.868}                                   & 30.36                                  & 0.9197                                  \\ \hline
			FSNet \cite{FSNet}                  &                                                                                           & 28.11                                  & 0.9056                                  & 34.80                                  & \multicolumn{1}{c|}{0.9067}                                  & 30.83                                  & 0.9686                                  & 28.29                                  & \multicolumn{1}{c|}{0.8305}                                  & 29.09                                  & 0.8991                                  \\
			FADC \cite{FADC}                    &                                                                                           & 29.49                                  & 0.9269                                  & 36.19                                  & \multicolumn{1}{c|}{0.9276}                                  & 32.10                                  & 0.9808                                  & 29.56                                  & \multicolumn{1}{c|}{0.8573}                                  & 30.03                                  & 0.9202                                  \\
			FDConv \cite{FDConv}                &                                                                                           & 28.86                                  & 0.9160                                  & 35.58                                  & \multicolumn{1}{c|}{0.9153}                                  & 31.51                                  & 0.9702                                  & 28.94                                  & \multicolumn{1}{c|}{0.8484}                                  & 29.55                                  & 0.9048                                  \\
			SFIN \cite{SFIN}                    &                                                                                           & {\ul 31.76}                            & {\ul 0.9543}                            & {\ul 38.48}                            & \multicolumn{1}{c|}{{\ul 0.9582}}                            & {\ul 33.24}                            & {\ul 0.9885}                            & {\ul 31.95}                            & \multicolumn{1}{c|}{{\ul 0.9028}}                            & {\ul 30.77}                            & {\ul 0.9333}                            \\
			\cellcolor[HTML]{F2F2F2}SFFN (Ours) & \multirow{-5}{*}{\begin{tabular}[c]{@{}c@{}}Spatial and Frequency\\ Network\end{tabular}} & \cellcolor[HTML]{F2F2F2}\textbf{32.83} & \cellcolor[HTML]{F2F2F2}\textbf{0.9649} & \cellcolor[HTML]{F2F2F2}\textbf{39.53} & \multicolumn{1}{c|}{\cellcolor[HTML]{F2F2F2}\textbf{0.9666}} & \cellcolor[HTML]{F2F2F2}\textbf{35.42} & \cellcolor[HTML]{F2F2F2}\textbf{0.9930} & \cellcolor[HTML]{F2F2F2}\textbf{32.74} & \multicolumn{1}{c|}{\cellcolor[HTML]{F2F2F2}\textbf{0.9214}} & \cellcolor[HTML]{F2F2F2}\textbf{32.32} & \cellcolor[HTML]{F2F2F2}\textbf{0.9540} \\ \hline
	\end{NiceTabular}}\label{tab:main}
	\vspace{0mm}
\end{table*}

\begin{figure*}
	\centering
	\includegraphics[width=\linewidth]{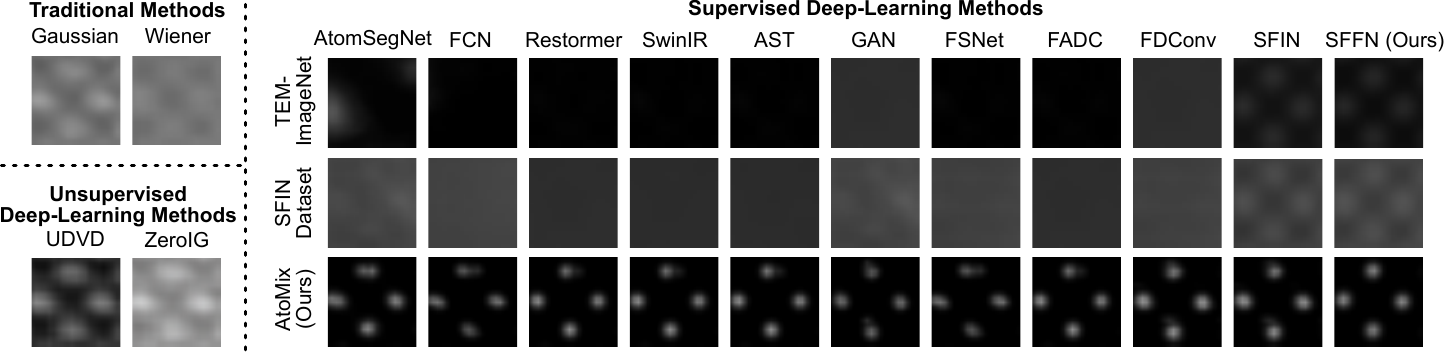}
	\vspace{-7mm}
	\caption{Results on a real STEM image sequence. The results on the right compare different networks trained on different datasets.}
	\label{fig:unsup}
	\vspace{-3mm}
\end{figure*}

\vspace{5pt}\noindent\textbf{Comparison methods.} 
Our Spatial-Frequency Fusion Network (SFFN) is compared with two traditional methods (\emph{i.e.}, Gaussian filter and Wiener filter \cite{Wiener}), ten supervised deep learning methods (\emph{i.e.}, AtomSegNet \cite{AtomSegNet}, FCN \cite{FCN}, Restormer \cite{Restormer}, SwinIR \cite{SwinIR}, AST \cite{AST}, GAN \cite{GAN}, FSNet \cite{FSNet}, FADC \cite{FADC}, FDConv \cite{FDConv}, SFIN \cite{SFIN}), and two unsupervised deep learning methods (\emph{i.e.}, UDVD \cite{UDVD} and ZeroIG \cite{ZeroIG}). AtomSegNet, FCN, Restormer, SwinIR, AST, and GAN are spatial domain networks, while FSNet, FADC, FDConv, SFIN, and our SFFN are spatial and frequency domain networks. Due to the requirement for multi-frame input images, UDVD and ZeroIG are only compared on the multi-frame real-world data captured in our laboratory.

\vspace{5pt}\noindent\textbf{Implementation details of our network.} 
Our SFFN is trained on AtoMix for 500 epochs using the Adam optimizer \cite{Adam} ($\beta_1$ = 0.9 and $\beta_2$ = 0.999) and an $L_1$ loss. For each of the HAADF and BF imaging modes, AtoMix contains 1,000 groups of training images and 100 groups of test images. More data can be easily synthesized if needed. The code is implemented in PyTorch \cite{PyTorch}. The initial learning rate is set to $2\times10^{-4}$ and reduced by half at epochs 250, 400, 450, and 475, following SFIN \cite{SFIN}. The mini-batch size is set to 8, allowing training to be completed on a single NVIDIA 3090 GPU.

\subsection{Main Results} 
We evaluate the effectiveness of our spatial-frequency fusion network on both synthetic datasets and real data. Specifically, the results on synthetic datasets focus on validating the effectiveness of our network architecture under unified training/evaluation protocols. The results on real data further compare the denoising effects of different combinations of network architectures and training datasets.

\vspace{2pt}\noindent\textbf{Results on synthetic datasets.} 
We compare our SFFN with existing methods on TEMImageNet \cite{AtomSegNet}, SFIN dataset \cite{SFIN}, and our AtoMix dataset. Fig.~\ref{fig:main} shows the visual comparisons. The first row presents the results on TEMImageNet. The outputs of Gaussian filter and Wiener filter \cite{Wiener} are blurred, while distinct artifacts are observed in the upper-left regions of the results generated by AtomSegNet \cite{AtomSegNet}, FCN \cite{FCN}, Restormer \cite{Restormer}, SwinIR \cite{SwinIR}, and AST \cite{AST}. In contrast, GAN \cite{GAN}, FSNet \cite{FSNet}, FADC \cite{FADC}, FDConv \cite{FDConv}, SFIN \cite{SFIN}, and our SFFN achieve clear enhancement results, among which the brightness distribution of our SFFN’s output is most consistent with the ground truth. The second row presents the results on the SFIN dataset. Only SFIN and our SFFN can effectively distinguish between two closely spaced atoms, while other methods tend to incorrectly merge them into one. The bottom three rows present the results on our AtoMix dataset. Only our SFFN recovers all atoms, including dark and small ones.

Table~\ref{tab:main} shows the quantitative results. It can be observed that the spatial and frequency-domain networks achieve better performance compared to pure spatial-domain networks, as they can exploit the periodicity of atomic arrangements in STEM images. Among the spatial and frequency-domain networks, our SFFN achieves the best performance across different datasets, with an approximate increase of 1 to 2 dB in PSNR compared to the second-best method SFIN \cite{SFIN}. This demonstrates the effectiveness of fusing information from the two domains.

\begin{figure}
	\centering
	\includegraphics[width=.9\linewidth]{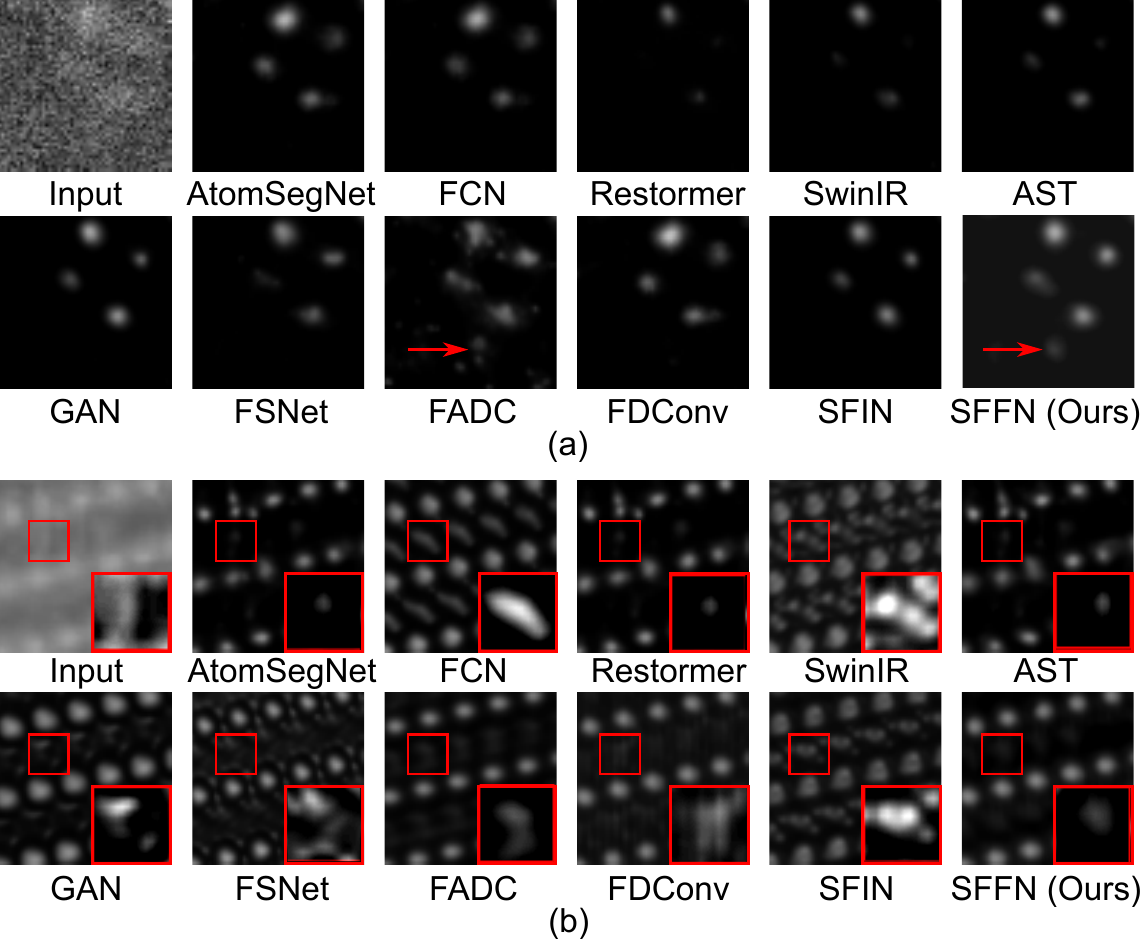}
	\vspace{-3mm}
	\caption{Comparison of different methods  on real data.}
	\label{fig:add_real}
\end{figure}

\begin{figure*}
	\centering
	\includegraphics[width=1\linewidth]{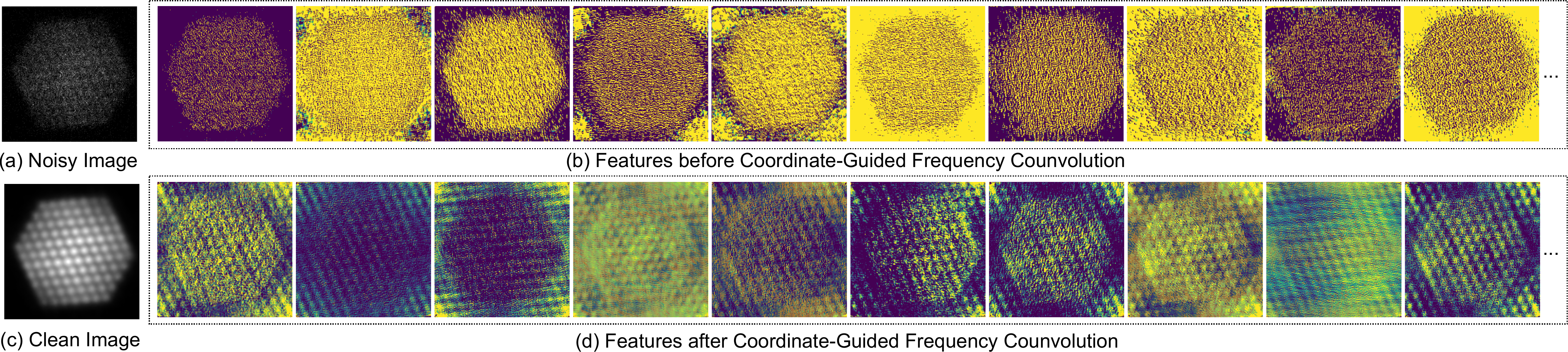}
	\vspace{-7mm}
	\caption{Features before and after coordinate-guided frequency convolution. The features shown in (b) and (d) come from the sixth coordinate-guided frequency convolution in our spatial-frequency fusion network. The periodic signals of atoms are significantly enhanced.}
	\vspace{-5mm}
	\label{fig:feature}
\end{figure*}

\begin{table}\small
	\centering
	\caption{Ablation results of our network. Models are trained and tested on our AtoMix dataset for HAADF mode image enhancement. Variants V1 to V5 feature different network structures, including whether spatial-domain and frequency-domain convolutions are used, how information interacts between the two domains, and whether coordinate information is employed in the coordinate-guided frequency convolutions. Best and second best values are \textbf{bold} and {\ul underlined}, respectively.}
	\vspace{-2mm}
	\resizebox{1.\columnwidth}{!}{
		\setlength{\tabcolsep}{1.mm}
		\begin{NiceTabular}{c|cccc|cc}
			\hline
			Method      & Spatial      & Frequency     & Interaction  & Coordinates  & PSNR $\uparrow$ & SSIM $\uparrow$ \\ \hline
			SFFN-V1     & \Checkmark   & \XSolidBrush & \XSolidBrush & -            & 27.80           & 0.8605          \\
			SFFN-V2     & \XSolidBrush & \Checkmark   & \XSolidBrush & \Checkmark   & 29.40           & 0.8880          \\
			SFFN-V3     & \Checkmark   & \Checkmark   & \XSolidBrush & \Checkmark   & 31.60           & 0.9060          \\
			SFFN-V4     & \Checkmark   & \Checkmark   & Addition     & \Checkmark   & 31.73           & 0.9072          \\
			SFFN-V5     & \Checkmark   & \Checkmark   & Fusion       & \XSolidBrush & {\ul 32.49}     & {\ul 0.9193}    \\
			\rowcolor[HTML]{F2F2F2} 
			SFFN (Ours) & \Checkmark   & \Checkmark   & Fusion       & \Checkmark   & \textbf{32.74}  & \textbf{0.9214} \\ \hline
	\end{NiceTabular}}\label{tab:ablation}
	\vspace{-2mm}
\end{table}

\vspace{2pt}\noindent\textbf{Results on real data.}
We collect a sequence of HAADF-mode images of a high-entropy alloy to comprehensively compare all methods, including traditional methods, supervised deep learning methods, and unsupervised deep learning methods.

The visual results are shown in Fig.~\ref{fig:unsup}. On the left part of Fig.~\ref{fig:unsup}, it can be seen that the atomic shapes in the enhanced results of Gaussian and Wiener filters \cite{Wiener}, UDVD \cite{UDVD}, and ZeroIG \cite{ZeroIG} are distorted (\emph{i.e.}, not a perfect circle). Meanwhile, the atomic sizes are relatively large, which is not conducive to analyzing atomic positions. This is because traditional and unsupervised methods are not trained with direct ground-truth constraints; although they can remove noise, they struggle to eliminate spherical aberration and background interference, as well as recover the shapes and sizes of atoms. 

On the right part of Fig.~\ref{fig:unsup} are the results of different supervised methods trained on different datasets. The atoms in the results of the last row are significantly clearer than those in the first two rows, indicating that our AtoMix dataset achieves better training effectiveness. In the last row, the atom located in the upper region exhibits artifacts or leftward positional deviations in the results of AtomSegNet \cite{AtomSegNet}, FCN \cite{FCN}, Restormer \cite{Restormer}, SwinIR \cite{SwinIR}, AST \cite{AST}, FSNet \cite{FSNet}, and FADC \cite{FADC}. These errors stem from the misclassification of noise as atomic contours. For GAN \cite{GAN}, FDConv \cite{FDConv}, and SFIN \cite{SFIN}, the upper atom is correctly recovered, yet a small artifact appears in the upper-left corner of the lower atom. In contrast, the result of our SFFN is free from artifacts, and the positions of all atoms are consistent with the material structure.

Fig.~\ref{fig:add_real} presents more complex real-world scenarios to compare different supervised learning methods. In Fig.~\ref{fig:add_real}(a), most methods fail to detect theoretically present atoms (\emph{e.g.}, at the red arrows). Although FADC recovers these atoms, it introduces significant shape distortions. Only our SFFN produces a complete and distortion-free image. In Fig.~\ref{fig:add_real}(b), only our SFFN yields undistorted atomic images. AtomSegNet, Restormer, and AST inaccurately recover bright atoms, while other competing methods exhibit noticeable distortions in dark atoms.

\subsection{Discussion}\label{sec:freq}
We first discuss the individual effectiveness of the core components of our SFFN, including coordinate-guided frequency convolution, frequency-coordinate embedding, and cross-domain feature fusion. Then, we discuss the effectiveness of our AtoMix dataset.

\begin{table}\small
	\centering
	\caption{STEM image enhancement results of different frequency-domain processing networks. Models are trained and tested on our dataset (HAADF mode). Best and second best values are \textbf{bold} and {\ul underlined}, respectively. Our SFFN achieves the best performance while tying SFIN for the fewest parameters.}
	\vspace{-2mm}
	\resizebox{\columnwidth}{!}{
		\setlength{\tabcolsep}{.4mm}
		\begin{NiceTabular}{
				>{\columncolor[HTML]{FFFFFF}}c |
				>{\columncolor[HTML]{FFFFFF}}c 
				>{\columncolor[HTML]{FFFFFF}}c 
				>{\columncolor[HTML]{FFFFFF}}c |
				>{\columncolor[HTML]{FFFFFF}}c 
				>{\columncolor[HTML]{FFFFFF}}c |
				>{\columncolor[HTML]{FFFFFF}}c 
				>{\columncolor[HTML]{FFFFFF}}c }
			\hline
			Method                 & \multicolumn{1}{l}{\cellcolor[HTML]{FFFFFF}\begin{tabular}[c]{@{}l@{}}Frequency-\\ position\\ guidance\end{tabular}} & \begin{tabular}[c]{@{}c@{}}Spatial-\\ frequency\\ interaction\end{tabular} & \begin{tabular}[c]{@{}c@{}}Per-frequency-\\ location\\ processing\end{tabular} & \begin{tabular}[c]{@{}c@{}}FLOPs\\ [G]\end{tabular} & \begin{tabular}[c]{@{}c@{}}Params\\ [M]\end{tabular} & PSNR $\uparrow$ & SSIM $\uparrow$ \\ \hline
			DeepRFT \cite{DeepRFT} & \XSolidBrush                                                                                                         & \begin{tabular}[c]{@{}c@{}}Fixed\\ (Addition)\end{tabular}                 & \Checkmark                                                                     & \textbf{45.7}                                       & 9.5                                                  & 30.62           & 0.8833          \\
			FSNet \cite{FSNet}     & \XSolidBrush                                                                                                         & \XSolidBrush                                                               & \XSolidBrush                                                                   & {\ul 59.6}                                          & 7                                                    & 28.29           & 0.8305          \\
			FADC \cite{FADC}       & \XSolidBrush                                                                                                         & \XSolidBrush                                                               & \XSolidBrush                                                                   & 62.5                                                & 1.2                                                  & 29.56           & 0.8573          \\
			FDConv \cite{FDConv}   & \XSolidBrush                                                                                                         & \XSolidBrush                                                               & \XSolidBrush                                                                   & 59.4                                                & {\ul 1.1}                                            & 28.94           & 0.8484          \\
			SFIN \cite{SFIN}       & \XSolidBrush                                                                                                         & \begin{tabular}[c]{@{}c@{}}Fixed\\ (Addition)\end{tabular}                 & \Checkmark                                                                     & 60.9                                                & \textbf{0.9}                                         & {\ul 31.95}     & {\ul 0.9028}    \\
			SFFN (Ours)            & \Checkmark                                                                                                           & Learnable                                                                  & \Checkmark                                                                     & 60.9                                                & \textbf{0.9}                                         & \textbf{32.74}  & \textbf{0.9214} \\ \hline
			\end{NiceTabular}}\label{tab:frequency}

\end{table}

\begin{figure}
	\centering
	\includegraphics[width=.9\linewidth]{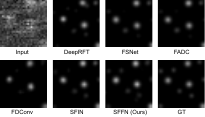}
	\vspace{-2mm}
	\caption{Comparison of frequency processing networks on AtoMix dataset.}
	\label{fig:frequency}

\end{figure}

\vspace{5pt}\noindent\textbf{Effectiveness of our coordinate-guided frequency convolution.}
We use coordinate-guided frequency convolution in each spatial-frequency fusion block to extract and enhance frequency-domain features. We first visualize its effect on enhancing periodic features and then compare it with other frequency-domain networks or modules.

Fig.~\ref{fig:feature} visualizes the enhancement effect of our coordinate-guided frequency convolution on the periodic information in STEM images. Fig.~\ref{fig:feature}(b) shows the feature map before enhancement, where the image signals are masked by noise, making it impossible to distinguish different atoms. After enhancement, each feature map in Fig.~\ref{fig:feature}(d) exhibits the same periodic variation as the real atomic arrangement in Fig.~\ref{fig:feature}(c). It indicates that our coordinate-guided frequency convolution can enhance the periodic information of atomic arrangements in the image, thereby helping to recover the atomic contours from the noise. Furthermore, the features in Fig.~\ref{fig:feature}(d) exhibit richer colors (such as blue and green) compared to those in Fig.~\ref{fig:feature}(b). This indicates that the values of the features have been equalized to distinguish different parts of each atom. Although Fig.~\ref{fig:feature}(d) exhibits clearer atomic periodicity, periodic responses also appear in the background regions. This indicates that frequency-domain processing alone cannot distinguish different spatial locations, motivating the fusion of frequency-domain and spatial-domain features. Table~\ref{tab:ablation} shows that, compared with the spatial-only variant SFFN-V1, introducing the coordinate-guided frequency branch in SFFN-V3 increases the PSNR from 27.80 dB to 31.60 dB, demonstrating the benefit of frequency-domain processing.

Table~\ref{tab:frequency} compares our coordinate-guided frequency convolution with other state-of-the-art frequency-domain networks or modules, including DeepRFT \cite{DeepRFT}, FSNet \cite{FSNet}, FADC \cite{FADC}, and FDConv \cite{FDConv}. Although these methods incorporate frequency-domain analysis, they fundamentally differ from our approach in three key aspects. First, instead of using frequency analysis merely to guide spatial convolutions (\emph{e.g.}, FADC) or performing coarse frequency selection (\emph{e.g.}, FSNet and FDConv), our method explicitly transforms features into the Fourier domain and applies convolution directly on frequency-domain features. This enables fine-grained modeling of distinct frequency locations, which is crucial for separating and enhancing individual diffraction spots in STEM images. Second, we introduce explicit frequency-position guidance by appending the x/y coordinates of the frequency map to the frequency-domain features. This makes the network explicitly aware of the positional context of each frequency component, enhancing its sensitivity to diffraction spot locations and thereby improving atomic structure recovery under heavy noise. Third, for explicit spatial-frequency feature fusion, rather than omitting cross-domain interaction (\emph{e.g.}, FSNet, FADC, FDConv) or relying on simple addition (\emph{e.g.}, SFIN and DeepRFT), we design parallel spatial and frequency branches fused via concatenation followed by a convolution. This learnable fusion provides greater flexibility than fixed addition and enables the network to combine complementary spatial- and frequency-domain information based on the input features. Consequently, the model can jointly leverage periodic structural priors from the frequency domain and local spatial anomalies, such as atomic vacancies, material interfaces, and embedded ions. As shown in Table~\ref{tab:frequency}, our SFFN achieves higher enhancement accuracy with only 0.9M parameters while maintaining computational complexity comparable to existing methods. Fig.~\ref{fig:frequency} shows that our frequency-domain processing method helps restore atomic structures more accurately.

\begin{figure}
	\centering
	\includegraphics[width=.8\linewidth]{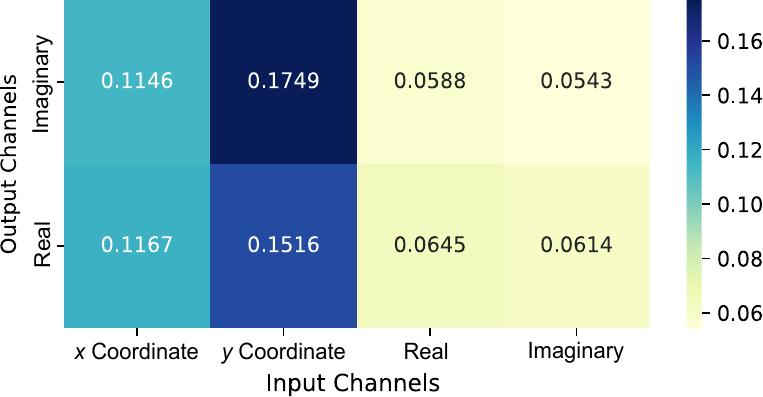}
	\vspace{-3mm}
	\caption{Importance of coordinate information in frequency-domain processing. The heat map shows the relative contributions of different input channels to the real and imaginary output channels of the 1$\times$1 convolution in CGFC. Each value is the mean absolute magnitude of the corresponding kernel weights across all 16 CGFC modules in SFFN. The $x$- and $y$-coordinate channels contribute substantially to frequency-domain processing.}
	\label{fig:conv_vis}
	\vspace{-2mm}
\end{figure}

\vspace{5pt}\noindent\textbf{Effectiveness of using frequency coordinates.}
Fig.~\ref{fig:conv_vis} shows the relative contributions of different input channels in the coordinate-guided frequency convolution, including the coordinate channels and the real and imaginary feature channels. Both coordinate channels contribute substantially to the real and imaginary outputs, indicating that frequency-position information helps the convolution distinguish different frequency locations and thereby separate signal from noise. The contribution of the $y$ coordinate is slightly larger than that of the $x$ coordinate because the height of the frequency-domain feature map is approximately twice its width (Fig.~\ref{fig:frame}(c)), allowing the $y$ coordinate to encode more detailed positional information. Table~\ref{tab:ablation} shows that using coordinate information (our SFFN), compared with omitting it (SFFN-V5), increases the PSNR by 0.25 dB.

\begin{figure}
	\centering
	\includegraphics[width=.60\linewidth]{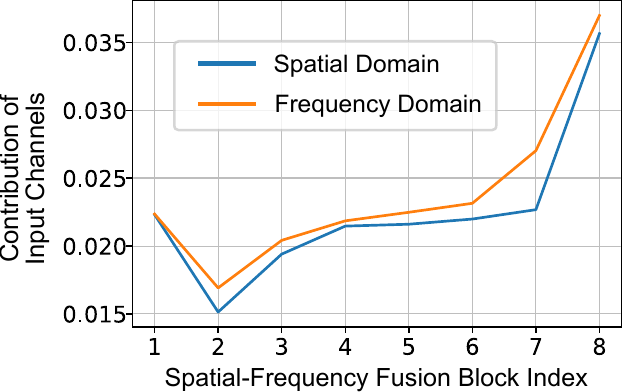}
	\vspace{-2mm}
	\caption{Importance of spatial- and frequency-domain features in cross-domain fusion. The curves show their relative contributions in each spatial-frequency fusion block, measured by the mean absolute kernel weights associated with the corresponding input channels in the final 3$\times$3 fusion convolution. Their similar magnitudes indicate that spatial- and frequency-domain features have comparable importance.}
	\label{fig:fusion_vis}
	\vspace{-2mm}
\end{figure}

\vspace{5pt}\noindent\textbf{Effectiveness of our cross-domain feature fusion.}
Fig.~\ref{fig:fusion_vis} compares the relative contributions of spatial- and frequency-domain features across different spatial-frequency fusion blocks. The two curves have similar magnitudes, indicating that both domains play important roles in STEM image enhancement. The contribution of frequency-domain features is consistently slightly greater than that of spatial-domain features, highlighting the importance of frequency-domain processing for periodic STEM images. In addition, the substantially larger mean absolute kernel weights in the final block indicate stronger cross-domain fusion near the network output.

The ablation results in Table~\ref{tab:ablation} indicate that compared with the simultaneous use of spatial-domain and frequency-domain information (SFFN-V3), the use of only spatial-domain information (SFFN-V1) or only frequency-domain information (SFFN-V2) both leads to performance degradation. This demonstrates that the information from both domains is important and complementary. Compared with no information interaction between the two domains (SFFN-V3), using addition for cross-domain information interaction (SFFN-V4) can increase the PSNR by 0.13 dB, while using the domain fusion operation in our SFFN can increase the PSNR by 1.14 dB. This indicates that domain fusion can more flexibly utilize information from both spatial and frequency domains to achieve better feature-enhancement results.

\begin{figure}
	\centering
	\includegraphics[width=1\linewidth]{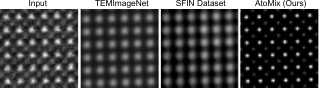}
	\vspace{-7mm}
	\caption{Result of our SFFN using different training datasets. The input image is real data, and the synthetic datasets used for training include TEMImageNet \cite{AtomSegNet}, SFIN dataset \cite{SFIN}, and our AtoMix dataset. Training with our AtoMix yields the best result.}
	\vspace{-3mm}
	\label{fig:data}
\end{figure}

\vspace{5pt}\noindent\textbf{Effectiveness of our dataset.} 
Our new dataset AtoMix adopts the noise parameters calibrated in SFIN \cite{SFIN}. Compared with the previous datasets where noise parameters were manually set, such as TEMImageNet \cite{AtomSegNet} and GAN dataset \cite{GAN-data}, the noise calibration makes the noise distribution in the simulated images closer to the real situation. The effectiveness of our noise calibration and simulation method has been verified in SFIN \cite{SFIN}. 

Table~\ref{tab:data} compares the differences among various datasets. Unlike the SFIN dataset \cite{SFIN}, our AtoMix dataset simulates more complex image degradations, including lower spatial resolution, lower atomic brightness, spherical aberration, and combinations of multiple arrangement patterns. These improvements enable our SFFN to handle more complex atomic arrangements and STEM image degradations.

The right part of Fig.~\ref{fig:unsup} demonstrates that the network trained on our AtoMix dataset better recovers atoms in real STEM images. Fig.~\ref{fig:data} further compares the results of our SFFN trained on different datasets. Training on AtoMix produces clear results for all atoms. In contrast, the results obtained using the other datasets are blurred and fail to recover relatively dark atoms.

\begin{table*}\small
	\centering
	\caption{Comparison of STEM image enhancement datasets. Our AtoMix is more diverse and includes more complex degradations.}
	\vspace{-2mm}
	\setlength{\tabcolsep}{.5pt}
	\resizebox{\textwidth}{!}{
		\setlength{\tabcolsep}{1.mm}
		\begin{NiceTabular}{c|cccc|cc|c|c|cc}
			\hline
			Dataset
			& \begin{tabular}[c]{@{}c@{}}Random\\ Scale\end{tabular}
			& \begin{tabular}[c]{@{}c@{}}Random\\ Rotation\end{tabular}
			& \begin{tabular}[c]{@{}c@{}}Random\\ Arrangements\end{tabular}
			& \begin{tabular}[c]{@{}c@{}}Mixed\\ Arrangements\end{tabular}
			& \begin{tabular}[c]{@{}c@{}}No. of\\ Patterns $\uparrow$\end{tabular}
			& \begin{tabular}[c]{@{}c@{}}No. of Crystal\\ Systems $\uparrow$\end{tabular}
			& \begin{tabular}[c]{@{}c@{}}BF\\ Mode\end{tabular}
			& \begin{tabular}[c]{@{}c@{}}Spherical\\ Aberration\end{tabular}
			& \begin{tabular}[c]{@{}c@{}}Minimum\\ Brightness $\downarrow$\end{tabular}
			& \begin{tabular}[c]{@{}c@{}}Minimum Atom\\ Diameter (pixels) $\downarrow$\end{tabular} \\ \hline
			TEMImageNet \cite{AtomSegNet} & Discrete     & Discrete        & \XSolidBrush       & \XSolidBrush    & 13                        & 4                         & \XSolidBrush & \Checkmark                                & 76                          & 14                          \\
			SFIN Dataset \cite{SFIN}      & Continuous   & Continuous      & \Checkmark         & \XSolidBrush    & 15                        & 5                         & \Checkmark   & \XSolidBrush                              & 119                         & 15                          \\
			\rowcolor[HTML]{F2F2F2} 
			AtoMix (Ours)                 & Continuous   & Continuous      & \Checkmark         & \Checkmark      & 46                        & 7                         & \Checkmark   & \Checkmark                                & 70                          & 7                           \\ \hline
			\end{NiceTabular}}\label{tab:data}
	\vspace{-2mm}
\end{table*}

\section{Application}
In this section, we evaluate the practical utility of our method, including its denoising performance on real images of advanced materials and its effectiveness in the downstream atomic localization task.

\subsection{Enhancement Results on Advanced Materials}\label{advanced} We evaluate the practical effectiveness of jointly using our SFFN architecture and AtoMix training dataset for observing advanced materials in real images. Specifically, we apply our Spatial-Frequency Fusion Network (SFFN) to images of superconducting materials, all-solid-state batteries, nanodevices, and neuromorphic computing materials. We compare it with representative STEM image enhancement methods, \emph{i.e.}, Gaussian filtering, Wiener filtering \cite{Wiener}, AtomSegNet \cite{AtomSegNet}, and SFIN \cite{SFIN}.

\begin{figure*}
	\centering
	\includegraphics[width=.9\linewidth]{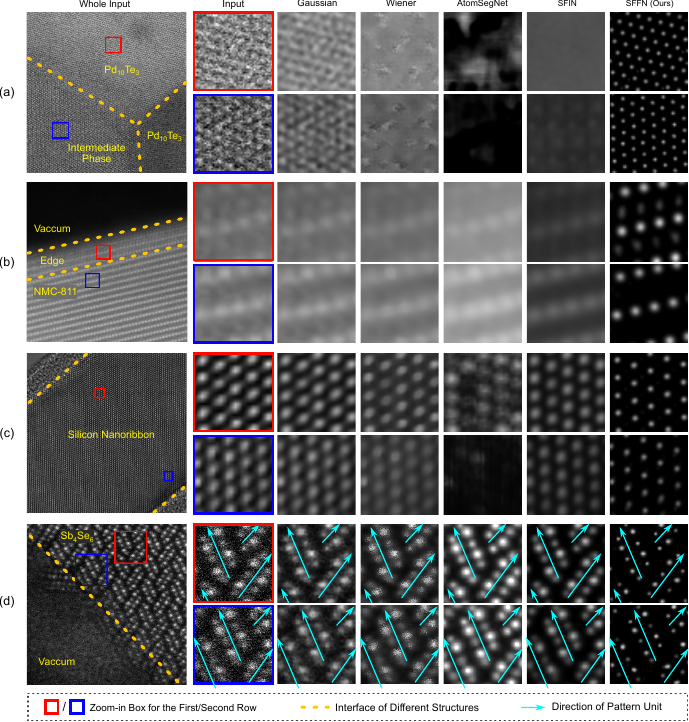}
	\vspace{-2mm}
	\caption{Enhancement results of real images. (a) Superconducting material \cite{superconduct}; (b) All-solid-state battery \cite{Battery1}; (c) Nanodevice \cite{heat}; (d) Neuromorphic computing material \cite{Interfaces}. Gaussian and Wiener \cite{Wiener} filters are traditional methods. AtomSegNet \cite{AtomSegNet}, SFIN \cite{SFIN}, and our SFFN are trained on their corresponding datasets. Our SFFN demonstrates better enhancement effects on these advanced materials.}
	\label{fig:real}
	\vspace{-5mm}
\end{figure*}

\vspace{2pt}\noindent\textbf{Superconducting material.} 
The nucleation of binary compounds into superconducting materials can be induced by altering their stoichiometry, and the analysis and control of this process depend on observation at the atomic scale \cite{superconduct}. We apply our enhancement method SFFN to observe the nucleation process from $\text{Pd}_{10}\text{Te}_{3}$ to $\text{Pd}\text{Te}_{2}$ \cite{superconduct}, as shown in Fig.~\ref{fig:real}(a). To ensure a large field of view, each atom in the image has a diameter of only 7 pixels, which poses significant challenges for image enhancement. The results of Gaussian filtering fail to distinguish different atoms, while those from Wiener filtering \cite{Wiener}, AtomSegNet \cite{AtomSegNet}, and SFIN \cite{SFIN} cannot even visualize atoms. In contrast, our SFFN clearly reveals the atomic structures of the nucleus and intermediate state, thanks to its more advanced network architecture and more diverse training data. It can be seen that compared with the perfectly regular $\text{Pd}_{10}\text{Te}_{3}$ nuclear structure, the intermediate state structure has atoms that are not completely aligned in the same straight line. These characterizations help to clearly demonstrate the atomic structures required for different superconducting properties.

\vspace{2pt}\noindent\textbf{All-solid-state battery.} 
The chemomechanical failure of layered cathodes in all-solid-state batteries affects the performance of the batteries, and its atomic origin has recently been revealed by STEM images \cite{Battery1}. We use our SFFN to characterize the structure of NMC-811 cathode after long-term contact with $\text{Li}_{6}\text{PS}_{5}\text{Cl}$ \cite{Battery1}, as shown in Fig.~\ref{fig:real}(b). Our results show that only slight cation mixing (showing some darker atoms caused by oxygen loss) occurred at the edge region. This corroborates the observations in the paper \cite{Battery1}. Our SFFN provides a method for evaluating the structure of all-solid-state batteries, which is difficult to achieve through direct observation of noisy images or the use of other enhancement methods \cite{Wiener,AtomSegNet,SFIN}.

\vspace{2pt}\noindent\textbf{Nanodevice.} 
Precise thermal management represents a critical bottleneck for the efficiency and service life of nanodevices \cite{heat}. We apply our SFFN to enhance high-resolution transmission electron microscopy (HRTEM) images of stressed and bent silicon nanoribbons \cite{heat}, as shown in Fig.~\ref{fig:real}(c). Unlike point-by-point scanned STEM images, HRTEM images are globally acquired and are therefore free from scan noise. The results demonstrate that our STEM image enhancement method is also applicable to other types of atomic-scale images. Specifically, although the input images exhibit no significant noise, atomic shape distortions arise from phase contrast and angular issues. Gaussian and Wiener filtering fail to remove the distortions, while AtomSegNet's enhancement results miss some atoms. SFIN can eliminate atomic distortions but remains blurry. Our SFFN clearly presents the central positions of atoms, which facilitates the measurement of distances between atoms and enables analysis of force at the atomic scale and the properties of nanodevices \cite{heat}.

\vspace{2pt}\noindent\textbf{Neuromorphic computing material.} 
The performance of neuromorphic computing requires an in-depth understanding of ion transport at material interfaces, which is often hindered by noise in STEM images \cite{Interfaces}. We apply our enhancement method SFFN to observe structural changes in $\text{Sb}_4\text{Se}_{6}$ during potassium-ion insertion \cite{Interfaces}, as shown in Fig.~\ref{fig:real}(d). The cyan arrows show that the atomic pattern units far from and near the interface (corresponding to the first and second rows, respectively) exhibit different arrangements, as indicated by the arrow in the lower-left corner. This difference arises because potassium-ion intercalation induces atomic rearrangement at the edge. Compared with other methods that suffer from spherical aberration, our SFFN removes spherical-aberration effects and produces more distinguishable atomic patterns.

\begin{figure}
	\centering
	\includegraphics[width=\linewidth]{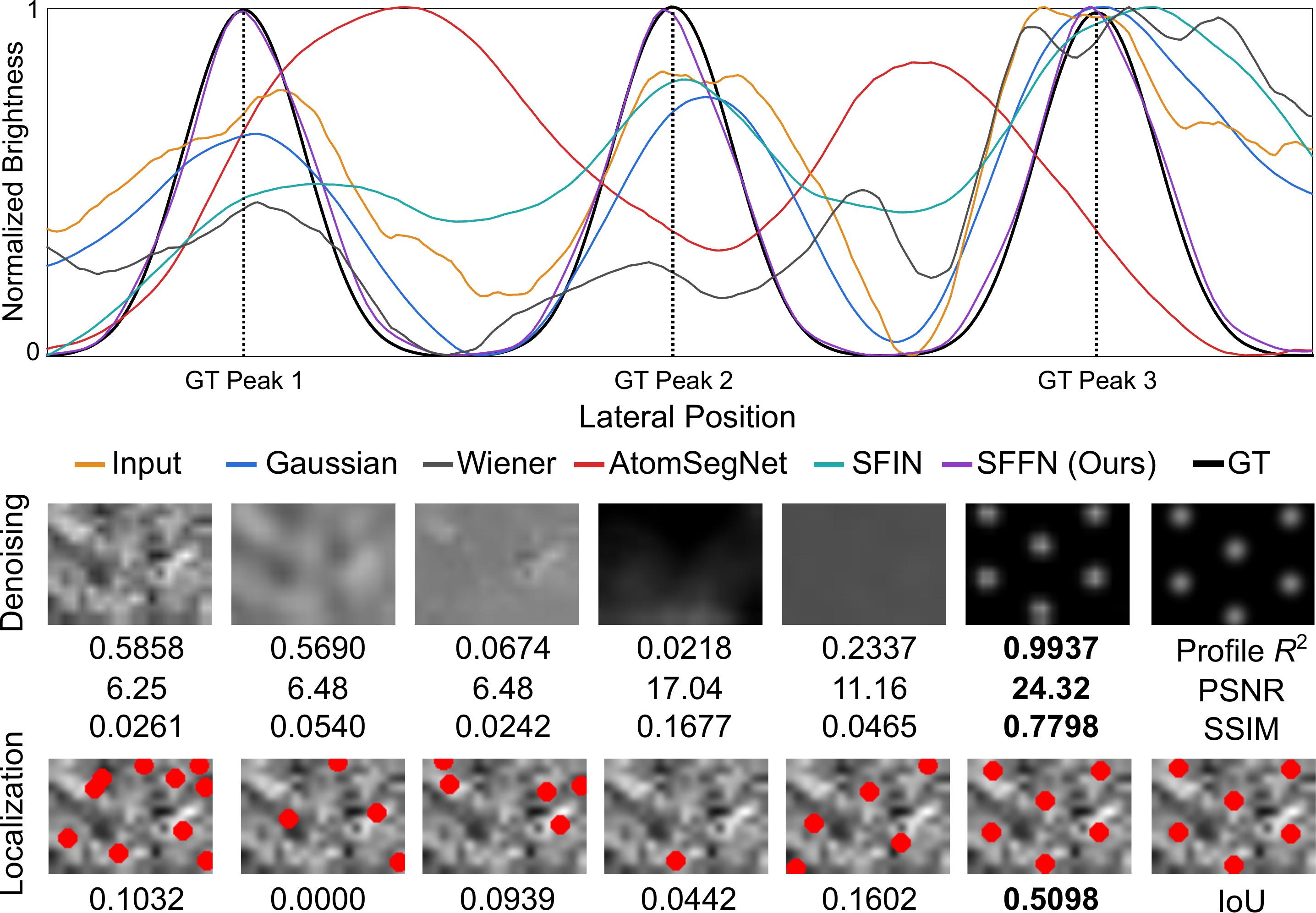}
	\vspace{-5mm}
	\caption{Quantitative analysis on a real image of $\text{Pd}_{10}\text{Te}_{3}$. GT is obtained through simulation and rendering of the crystal structure. (a) Line profile extracted from a narrow strip region. Our denoising results exhibit the closest brightness variation curve to the ground truth. (b) Denoising results. (c) Atomic localization results.}
	\label{fig:line-profiles}
	\vspace{-5mm}
\end{figure}

\vspace{2pt}\noindent\textbf{Quantitative analysis.} We further quantitatively compare different methods using a real image with a known crystal structure, as shown in Fig.~\ref{fig:line-profiles}. The input is an experimentally acquired image of $\text{Pd}_{10}\text{Te}_{3}$. Since the crystal structure is known, we obtain the corresponding ground-truth image through atomistic simulation and rendering. We determine the lattice periodicity and phase using line-profile and FFT analyses, and align the lattice arrangement of the ground-truth image with that of the input image. In Fig.~\ref{fig:line-profiles}(a), we compare the normalized brightness profiles extracted by averaging pixel intensities within the same narrow strip region. The line profile of our SFFN is the closest to the ground truth, indicating that our method better preserves the brightness variation and peak positions of atomic columns while suppressing noise and background interference. In Fig.~\ref{fig:line-profiles}(b), we further report the coefficient of determination ($R^2$) between the intensity profile of each denoising result and that of the ground truth. Our SFFN achieves the highest $R^2$ value of 0.9937, quantitatively demonstrating that its denoising result is most consistent with the simulated crystal structure. Our SFFN also achieves the highest PSNR of 24.32 dB and SSIM of 0.7798. In Fig.~\ref{fig:line-profiles}(c), we evaluate the atomic localization results using IoU. Our SFFN obtains the highest IoU of 0.5098, showing that the enhanced image produced by our method better supports downstream atomic localization.

\begin{figure*}
	\centering
	\includegraphics[width=.9\linewidth]{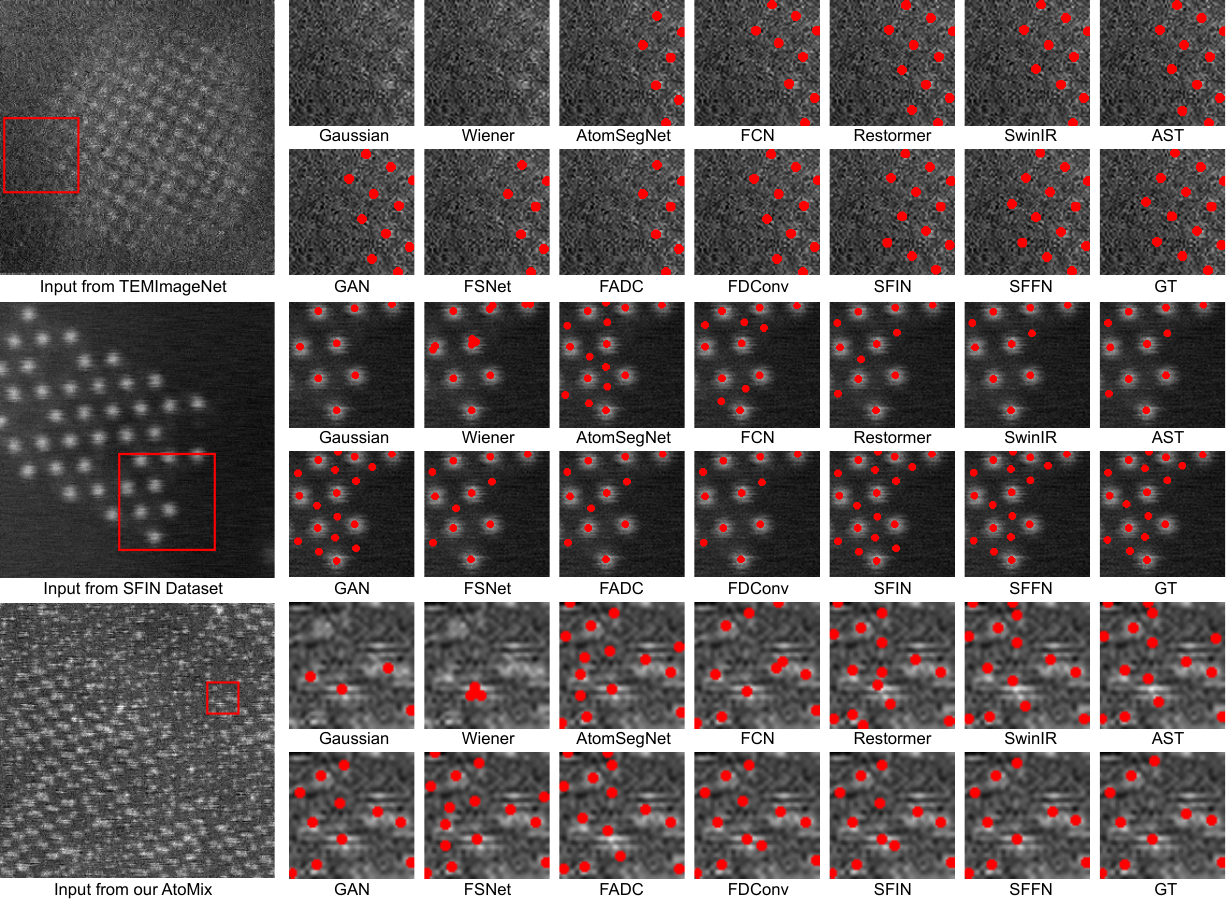}
	\vspace{-4mm}
	\caption{Localization results on different synthetic datasets. The training data used corresponds to the training set associated with the test data.}
	\label{fig:local_atomix}
	\vspace{0mm}
\end{figure*}

\begin{table*}\small
	\centering
	\caption{Localization results (IoU) on different synthetic datasets. Deep learning methods are trained and tested on the corresponding datasets. HAADF and BF are imaging modes.}
	\vspace{-2mm}
	\resizebox{.8\textwidth}{!}{
		\setlength{\tabcolsep}{1.mm}
		\begin{NiceTabular}{c|c|c|cc|cc}
			\hline
			&                                                                                           & TEMImageNet \cite{AtomSegNet}           & \multicolumn{2}{c|}{SFIN Dataset \cite{SFIN}}                                     & \multicolumn{2}{c}{AtoMix (Ours)}                                                 \\ \cline{3-7} 
			\multirow{-2}{*}{Method}            & \multirow{-2}{*}{Category}                                                                & HAADF Mode                              & HAADF Mode                              & BF Mode                                 & HAADF Mode                              & BF Mode                                 \\ \hline
			Gaussian Filter \cite{Bilateral}    &                                                                                           & 0.1251                                  & 0.0965                                  & 0.0443                                  & 0.1655                                  & 0.1804                                  \\
			Wiener Filter \cite{Wiener}         & \multirow{-2}{*}{Traditional}                                                             & 0.1144                                  & 0.0896                                  & 0.0439                                  & 0.1238                                  & 0.1397                                  \\ \hline
			AtomSegNet \cite{AtomSegNet}        &                                                                                           & 0.5636                                  & 0.4507                                  & 0.3581                                  & 0.3601                                  & 0.3902                                  \\
			FCN \cite{FCN}                      &                                                                                           & 0.6061                                  & 0.4822                                  & 0.4096                                  & 0.2880                                  & 0.3451                                  \\
			Restormer \cite{Restormer}          &                                                                                           & 0.6650                                  & 0.5206                                  & 0.4513                                  & 0.4104                                  & 0.4001                                  \\
			SwinIR \cite{SwinIR}                &                                                                                           & 0.6523                                  & 0.5150                                  & 0.4411                                  & 0.4084                                  & 0.4120                                  \\
			AST \cite{AST}                      &                                                                                           & 0.6462                                  & 0.5035                                  & 0.4369                                  & 0.3325                                  & 0.3664                                  \\
			GAN \cite{GAN}                      & \multirow{-6}{*}{\begin{tabular}[c]{@{}c@{}}Spatial\\ Network\end{tabular}}               & {\ul 0.7089}                            & 0.5496                                  & 0.4481                                  & 0.4208                                  & 0.4778                                  \\ \hline
			FSNet \cite{FSNet}                  &                                                                                           & 0.6511                                  & 0.5187                                  & 0.4329                                  & 0.4671                                  & 0.4147                                  \\
			FADC \cite{FADC}                    &                                                                                           & 0.6922                                  & 0.5428                                  & 0.4701                                  & 0.4398                                  & 0.4846                                  \\
			FDConv \cite{FDConv}                &                                                                                           & 0.6766                                  & 0.5271                                  & 0.4377                                  & 0.4239                                  & 0.4382                                  \\
			SFIN \cite{SFIN}                    &                                                                                           & 0.5987                                  & {\ul 0.5694}                            & {\ul 0.4814}                            & {\ul 0.4809}                            & {\ul 0.5236}                            \\
			\cellcolor[HTML]{F2F2F2}SFFN (Ours) & \multirow{-5}{*}{\begin{tabular}[c]{@{}c@{}}Spatial and Frequency\\ Network\end{tabular}} & \cellcolor[HTML]{F2F2F2}\textbf{0.7765} & \cellcolor[HTML]{F2F2F2}\textbf{0.5950} & \cellcolor[HTML]{F2F2F2}\textbf{0.5224} & \cellcolor[HTML]{F2F2F2}\textbf{0.5586} & \cellcolor[HTML]{F2F2F2}\textbf{0.5911} \\ \hline
	\end{NiceTabular}}\label{tab:loc}
	\vspace{-2mm}
\end{table*}

\subsection{Localization Results}
We compare the influence of different enhancement network architectures on downstream atomic localization under the same localization protocol.

After a noisy image is enhanced, atomic positions can be automatically detected to analyze the material's atomic arrangement \cite{AtomAI,AtomSegNet}. Following the protocols of AtomSegNet \cite{AtomSegNet} and SFIN \cite{SFIN}, we compare the effects of different enhancement methods on atomic localization. Specifically, we train each enhancement network using data designed for atomic localization, in which atoms of varying sizes and brightness are represented by small white disks so that each atom can be identified clearly. The network output is then binarized into a mask. The geometric center of each connected white region gives an atomic position.

Fig.~\ref{fig:local_atomix} presents the atomic localization results of different methods on various datasets. Gaussian filter and Wiener filter \cite{Wiener} fail to correctly identify most atoms. Among existing deep learning-based methods, SFIN \cite{SFIN} achieves relatively satisfactory atomic localization performance but still suffers from missed detections (\emph{e.g.}, its results on TEMImageNet \cite{AtomSegNet}) and false detections (\emph{e.g.}, its results on the SFIN dataset \cite{SFIN} and our AtoMix). Our SFFN fuses and leverages both frequency-domain and spatial-domain features, which facilitates distinguishing weak signals from noise, thereby yielding results consistent with the ground truth. Table~\ref{tab:loc} also demonstrates that our SFFN achieves the best Intersection over Union (IoU) results on the three atomic localization datasets. 

\begin{figure*}
	\centering
	\includegraphics[width=\linewidth]{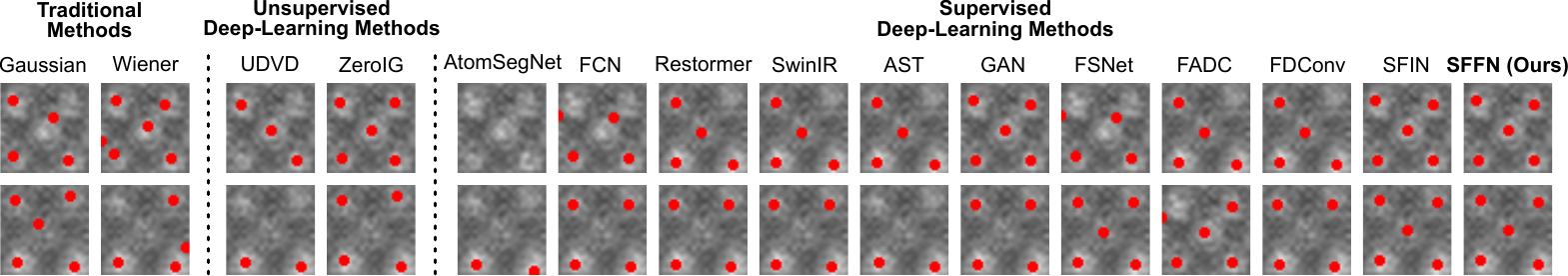}
	\vspace{-7mm}
	\caption{Localization results on real data. All the supervised deep-learning methods are trained on our AtoMix.}
	\label{fig:unsup-local}
	\vspace{-3mm}
\end{figure*}

Fig.~\ref{fig:unsup-local} shows the atomic localization results of traditional methods, unsupervised learning methods, and supervised learning methods on a sequence of real STEM images, where the material is a high-entropy alloy. SFIN \cite{SFIN} and our SFFN achieve the most complete localization results, which are consistent with the theoretical structure of the material.

\vspace{-1mm}
\section{Conclusion}
\vspace{-1mm}
Noise and aberrations in STEM images hinder in situ atomic-scale observation of new-material structures. We present a spatial-frequency fusion network and develop a realistic and diverse AtoMix dataset for STEM image enhancement. Our spatial-frequency fusion network exploits frequency-domain information from atomic periodicity and enhances and fuses spatial- and frequency-domain features. Specifically, the coordinate-guided frequency convolution processes features according to their frequency positions, thereby enhancing periodic signals. Furthermore, we develop a STEM noise calibration method that derives noise parameters from real images to synthesize realistic STEM images. AtoMix contains diverse atomic arrangements, imaging modes, and degradations and uses masks to simulate multimaterial interfaces. This diversity improves robustness in real-world STEM image applications. Experiments demonstrate that our method achieves state-of-the-art performance on both synthetic datasets and real images of advanced materials, supporting STEM-based materials research.

The experimental results demonstrate the significant enhancement effect of frequency-domain processing on STEM images. Since atomic periodic signals and noise have distinct distributions in the frequency domain, future research can combine frequency-domain processing with attention mechanisms, dynamic convolutions, \emph{etc.}, enabling the network to distinguish and treat effective information and noise differently in the frequency domain. In addition, it is worthwhile to extend our method to higher-dimensional data, such as 4D-STEM data, which provide a two-dimensional diffraction pattern at each two-dimensional probe position and are important for analyzing local structures, strain, and electromagnetic fields.

\bibliographystyle{IEEEtran}
\bibliography{reference.bib}

@String(PAMI = {IEEE Trans. Pattern Anal. Mach. Intell.})

@String(IJCV = {Int. J. Comput. Vis.})

@String(CVPR= {Proc. IEEE Conf. Comput. Vis. Pattern Recog.})

@String(ICCV= {Proc. Int. Conf. Comput. Vis.})

@String(NIPS= {Proc. Adv. Neural Inform. Process. Syst.})

@String(TIP  = {IEEE Trans. Image Process.})

@String(TMM  = {IEEE Trans. Multimedia})

@String(AAAI = {Proc. AAAI})

@String(CVPRW= {Proc. IEEE Conf. Comput. Vis. Pattern Recog. Worksh.})

@String(CSVT = {IEEE Trans. Circuit Syst. Video Technol.})

@String(CVM = {Comput. Vis. Media})

@String(Nature = {Nature})

@String(NatCommun = {Nat. Commun.})

@String(JACS = {J. Am. Chem. Soc.})

@String(NatMater = {Nat. Mater.})

@String(NatEnergy = {Nat. Energy})

@String(Science = {Science})

@String(OpenResEur = {Open Res. Eur.})

@String(npjCompMat = {npj Comput. Mater.})

@String(NanoLett = {Nano Lett.})

@String(SciRep = {Sci. Rep.})

@String(Elife = {eLife})

@String(MicroscMicroanal = {Microsc. Microanal.})

@String(Ultramicroscopy = {Ultramicrosc.})

@String(Springer = {Springer Berlin Heidelberg})

@String(Siggraph = {Proc. Conf. Comput. Graph. Interact. Tech.})

@String(NatMachIntell = {Nat. Mach. Intell.})

@String(JImgGraph = {J. Image Graph.})

@String(CJE = {Chin. J. Electron.})

@String(arXiv = {arXiv preprint})

@String(MMVC = {Proc. Med. Image Comput. Comput.-Assisted Interv.})

@String(TsinghuaSciTechnol = {Tsinghua Sci. Technol.})

@String(SpaceSciTechnol = {Space Sci. Technol.})

@String(ArtifIntell = {Artif. Intell.})

@article{semiconductor,
	author = {Sánchez-Santolino, G. and Rouco, V. and Puebla, S. and Aramberri, H. and Zamora, V. and Cabero, M. and Cuellar, F. A. and Munuera, C. and Mompean, F. and Garcia-Hernandez, M. and Castellanos-Gomez, A. and Íñiguez, J. and Leon, C. and Santamaria, J.},
	title = {A 2D ferroelectric vortex pattern in twisted BaTiO3 freestanding layers},
	journal = Nature,
	volume = {626},
	number = {7999},
	pages = {529--534},
	year = {2024},
}

@article{semiconductor2,
	title = {Polar Bloch points in strained ferroelectric films},
	volume = {15},
	pages = {3949},
	number = {1},
	journal = NatCommun,
	author = {Wang, Yu-Jia and Feng, Yan-Peng and Tang, Yun-Long and Zhu, Yin-Lian and Cao, Yi and Zou, Min-Jie and Geng, Wan-Rong and Ma, Xiu-Liang},
	year = {2024},
}

@article{Battery1,
	author = {Wang, Chunyang and Jing, Yaqi and Zhu, Dong and Xin, Huolin L.},
	title = {Atomic Origin of Chemomechanical Failure of Layered Cathodes in All-Solid-State Batteries},
	journal = JACS,
	volume = {146},
	number = {26},
	pages = {17712--17718},
	year = {2024},
}

@article{Battery2,
	title = {Resolving complex intralayer transition motifs in high-Ni-content layered cathode materials for lithium-ion batteries},
	author = {Wang, Chunyang and Wang, Xuelong and Zhang, Rui and Lei, Tianjiao and Kisslinger, Kim and Xin, Huolin L.},
	journal = NatMater,
	number = {2},
	volume = {22},
	pages = {235--241},
	year = {2023},
}

@article{Battery3,
	title = {Long-life lithium-ion batteries realized by low-Ni, Co-free cathode chemistry},
	author = {Zhang, Rui and Wang, Chunyang and Zou, Peichao and Lin, Ruoqian and Ma, Lu and Li, Tianyi and Hwang, In-hui and Xu, Wenqian and Sun, Chengjun and Trask, Steve and Xin, Huolin L.},
	journal = NatEnergy,
	pages = {695--702},
	number = {7},
	volume = {8},
	year = {2023},
}

@article{functional,
	title = {Chemical inhomogeneity–induced profuse nanotwinning and phase transformation in AuCu nanowires},
	author = {Yang, Chengpeng and Zhang, Bozhao and Fu, Libo and Wang, Zhanxin and Teng, Jiao and Shao, Ruiwen and Wu, Ziqi and Chang, Xiaoxue and Ding, Jun and Wang, Lihua and Han, Xiaodong},
	journal = NatCommun,
	number = {1},
	pages = {5705},
	volume = {14},
	year = {2023},
}

@article{functional2,
	title = {Insulating electromagnetic-shielding silicone compound enables direct potting electronics},
	volume = {385},
	pages = {1205--1210},
	number = {6714},
	journal = Science,
	year = {2024},
	author = {Zhou, Xinfeng and Min, Peng and Liu, Yue and Jin, Meng and Yu, Zhong-Zhen and Zhang, Hao-Bin}
}

@article{abTEM,
	author = {Madsen, Jacob and Susi, Toma},
	year = {2021},
	pages = {24},
	title = {The {abTEM} code: transmission electron microscopy from first principles},
	volume = {1},
	journal = OpenResEur,
}

@article{GAN-data,
	title={Leveraging generative adversarial networks to create realistic scanning transmission electron microscopy images},
	author={Khan, Abid and Lee, Chia-Hao and Huang, Pinshane Y and Clark, Bryan K},
	journal = npjCompMat,
	volume = {9},
	number = {1},
	pages = {85},
	year = {2023}
}

@article{FCN,
	title={Deep learning enabled strain mapping of single-atom defects in two-dimensional transition metal dichalcogenides with sub-picometer precision},
	author={Lee, Chia-Hao and Khan, Abid and Luo, Di and Santos, Tatiane P and Shi, Chuqiao and Janicek, Blanka E and Kang, Sangmin and Zhu, Wenjuan and Sobh, Nahil A and Schleife, Andr{\'e} and others},
	journal = NanoLett,
	volume = {20},
	number = {5},
	pages = {3369--3377},
	year = {2020}
}

@article{AtomSegNet,
	title={TEMImageNet training library and AtomSegNet deep-learning models for high-precision atom segmentation, localization, denoising, and deblurring of atomic-resolution images},
	author={Lin, Ruoqian and Zhang, Rui and Wang, Chunyang and Yang, Xiao-Qing and Xin, Huolin L},
	journal = SciRep,
	volume = {11},
	number = {1},
	pages = {5386},
	year = {2021}
}

@inproceedings{calib1,
	title={A physics-based noise formation model for extreme low-light raw denoising},
	author={Wei, Kaixuan and Fu, Ying and Yang, Jiaolong and Huang, Hua},
	booktitle = CVPR,
	pages = {2758--2767},
	year = {2020}
}

@article{calib2,
	title={Physics-based noise modeling for extreme low-light photography},
	author={Wei, Kaixuan and Fu, Ying and Zheng, Yinqiang and Yang, Jiaolong},
	journal = PAMI,
	volume = {44},
	number = {11},
	pages = {8520--8537},
	year = {2021}
}

@inproceedings{GAN,
	title={Self-Supervised Learning with Generative Adversarial Networks for Electron Microscopy},
	author={Kazimi, Bashir and Ruzaeva, Karina and Sandfeld, Stefan},
	booktitle = CVPRW,
	pages = {71--81},
	year = {2024}
}

@article{CEM500k,
	title={CEM500K, a large-scale heterogeneous unlabeled cellular electron microscopy image dataset for deep learning},
	author={Conrad, Ryan and Narayan, Kedar},
	journal = Elife,
	volume = {10},
	pages = {e65894},
	year = {2021}
}

@book{Wiener,
	title={Extrapolation, interpolation, and smoothing of stationary time series: with engineering applications},
	author={Wiener, Norbert},
	year={1949},
	publisher={The MIT press}
}

@inproceedings{Bilateral,
	title={Bilateral filtering for gray and color images},
	author={Tomasi, Carlo and Manduchi, Roberto},
	booktitle = ICCV,
	pages = {839--846},
	year = {1998},
}

@inproceedings{UNet,
	title={U-net: Convolutional networks for biomedical image segmentation},
	author={Ronneberger, Olaf and Fischer, Philipp and Brox, Thomas},
	booktitle = MMVC,
	pages = {234--241},
	year = {2015}
}

@article{PyTorch,
	title={{PyTorch}: An Imperative Style, High-Performance Deep Learning Library},
	author={Paszke, Adam and Gross, Sam and Massa, Francisco and Lerer, Adam and Chintala, Soumith},
	year={2019},
	journal = arXiv,
	note={arXiv:1912.01703}
}

@article{Adam,
	title={Adam: A method for stochastic optimization},
	author={Kingma, Diederik P and Ba, Jimmy},
	journal = arXiv,
	note={arXiv:1412.6980},
	year={2014}
}

@article{SSIM,
	title={Image quality assessment: from error visibility to structural similarity},
	author={Wang, Zhou and Bovik, Alan C and Sheikh, Hamid R and Simoncelli, Eero P},
	journal = TIP,
	volume = {13},
	number = {4},
	pages = {600--612},
	year = {2004},
}

@article{bf1,
	author = {Pennycook, SJ and Jesson, DE and Chisholm, MF and Browning, ND and McGibbon, AJ and McGibbon, MM},
	title = {Z-Contrast Imaging in the Scanning Transmission Electron Microscope},
	journal = MicroscMicroanal,
	volume = {1},
	number = {6},
	pages = {231--251},
	year = {2003},
}

@article{bf2,
	title = {High-resolution Z-contrast imaging of crystals},
	journal = Ultramicroscopy,
	volume = {37},
	number = {1},
	pages = {14--38},
	year = {1991},
	author = {S.J. Pennycook and D.E. Jesson},
}

@Inbook{bf3,
	author={Fultz, Brent and Howe, James},
	title={Diffraction Contrast in TEM Images},
	chapter={Transmission Electron Microscopy and Diffractometry of Materials},
	year={2013},
	pages={349--427},
	publisher = Springer,
}

@inproceedings{Perlin,
	author = {Perlin, Ken},
	title = {An image synthesizer},
	year = {1985},
	booktitle = Siggraph,
	pages = {287--296},
}

@INPROCEEDINGS{AST,
	author={Zhou, Shihao and Chen, Duosheng and Pan, Jinshan and Shi, Jinglei and Yang, Jufeng},
	booktitle = CVPR,
	title={Adapt or Perish: Adaptive Sparse Transformer with Attentive Feature Refinement for Image Restoration},
	year={2024},
	pages={2952--2963},
}

@ARTICLE{FSNet,
	author={Cui, Yuning and Ren, Wenqi and Cao, Xiaochun and Knoll, Alois},
	journal = PAMI,
	title={Image Restoration via Frequency Selection},
	year={2024},
	volume={46},
	number={2},
	pages={1093--1108},
}

@article{AtomAI,
	title={AtomAI framework for deep learning analysis of image and spectroscopy data in electron and scanning probe microscopy},
	author={Ziatdinov, Maxim and Ghosh, Ayana and Wong, Chun Yin and Kalinin, Sergei V},
	journal = NatMachIntell,
	volume = {4},
	number = {12},
	pages = {1101--1112},
	year = {2022},
}

@article{FCDFusion,
	title={{FCDFusion}: A fast, low color deviation method for fusing visible and infrared image pairs},
	author={Hesong Li and Ying Fu},
	journal = CVM,
	volume = {11},
	number = {1},
	pages = {195--211},
	year = {2025},
}

@article{JIG,
	author    = {Miaoyu Li and Ying Fu},
	title     = {Joint self-attention Transformer for multispectral and hyperspectral image fusion},
	journal   = JImgGraph,
	year      = {2023},
	volume    = {28},
	number    = {12},
	pages     = {3922--3934},
}

@article{CJE1,
	title={Non-uniform Compressive Sensing Imaging Based on Image Saliency},
	author={Li, Hongliang and Dai, Feng and Zhao, Qiang and Ma, Yike and Cao, Juan and Zhang, Yongdong},
	journal = CJE,
	volume = {32},
	number = {1},
	pages = {159--165},
	year = {2023}
}

@Article{CJE2,
	title = {Hyperspectral Image Super-Resolution Based on Spatial-Spectral Feature Extraction Network},
	journal = CJE,
	volume = {32},
	number = {3},
	pages = {415--428},
	year = {2023},
	author = {LI Yanshan and CHEN Shifu and LUO Wenhan and ZHOU Li and XIE Weixin}
}

@article{CJE3,
	title={Transformer-based under-sampled single-pixel imaging},
	author={Tian, Ye and Fu, Ying and Zhang, Jun},
	journal = CJE,
	volume = {32},
	number = {5},
	pages = {1151--1159},
	year = {2023}
}

@article{CJE4,
	title = {Deep Guided Attention Network for Joint Denoising and Demosaicing in Real Image},
	author = {Zhang, Tao and Fu, Ying and Zhang, Jun and Yan, Chenggang},
	journal = CJE,
	volume = {33},
	number = {1},
	pages = {303--312},
	year = {2024},
}

@article{he1,
	title={Diffusion Models in Low-Level Vision: A Survey},
	author={He, Chunming and Shen, Yuqi and Fang, Chengyu and Xiao, Fengyang and Tang, Longxiang and Zhang, Yulun and Zuo, Wangmeng and Guo, Zhenhua and Li, Xiu},
	journal = PAMI,
	year={2025},
	volume = {47},
	number = {6},
	pages = {4630--4651}
}

@article{he2,
	title={Segment concealed object with incomplete supervision},
	author={He, Chunming and Li, Kai and Zhang, Yachao and Yang, Ziyun and Tang, Longxiang and Zhang, Yulun and Kong, Linghe and Farsiu, Sina},
	journal = PAMI,
	year={2025},
	pages = {1--18}
}

@InProceedings{SFIN,
	author = {Li, Hesong and Wu, Ziqi and Shao, Ruiwen and Zhang, Tao and Fu, Ying},
	title = {Noise Calibration and Spatial-Frequency Interactive Network for {STEM} Image Enhancement},
	booktitle = CVPR,
	month = {June},
	year = {2025},
	pages = {21287--21296}
}

@article{superconduct,
	title={Stoichiometry-engineered phase transition in a two-dimensional binary compound},
	author={Huang, Mengting and Hua, Ze and Guzman, Roger and Ren, Zhihui and Gu, Pingfan and Yang, Shiqi and Chen, Hui and Zhang, Decheng and Ding, Yiming and Ye, Yu and others},
	journal = NatCommun,
	volume = {16},
	number = {1},
	pages = {4162},
	year = {2025},
	publisher={Nature Publishing Group UK London}
}

@article{heat,
	title={Suppressed thermal transport in silicon nanoribbons by inhomogeneous strain},
	author={Yang, Lin and Yue, Shengying and Tao, Yi and Qiao, Shuo and Li, Hang and Dai, Zhaohe and Song, Bai and Chen, Yunfei and Du, Jinlong and Li, Deyu and others},
	journal = Nature,
	volume = {629},
	number = {8014},
	pages = {1021--1026},
	year = {2024},
}

@InProceedings{Restormer,
	author    = {Zamir, Syed Waqas and Arora, Aditya and Khan, Salman and Hayat, Munawar and Khan, Fahad Shahbaz and Yang, Ming-Hsuan},
	title     = {Restormer: Efficient Transformer for High-Resolution Image Restoration},
	booktitle = CVPR,
	year      = {2022},
	pages     = {5728--5739}
}

@INPROCEEDINGS{SwinIR,
	author={Liang, Jingyun and Cao, Jiezhang and Sun, Guolei and Zhang, Kai and Van Gool, Luc and Timofte, Radu},
	booktitle = {IEEE Int. Conf. Comput. Vis. Worksh.},
	title={SwinIR: Image Restoration Using Swin Transformer},
	year={2021},
	pages={1833--1844}
}

@inproceedings{FADC,
	title={Frequency-Adaptive Dilated Convolution for Semantic Segmentation},
	author={Chen, Linwei and Gu, Lin and Zheng, Dezhi and Fu, Ying},
	booktitle = CVPR,
	pages = {3414--3425},
	year = {2024}
}

@InProceedings{FDConv,
	title={Frequency Dynamic Convolution for Dense Image Prediction},
	author={Chen, Linwei and Gu, Lin and Li, Liang and Yan, Chenggang and Fu, Ying},
	booktitle = CVPR,
	pages={30178--30188},
	year={2025}
}

@article{UDVD,
	title={Visualizing nanoparticle surface dynamics and instabilities enabled by deep denoising},
	author={Crozier, Peter A and Leibovich, Matan and Haluai, Piyush and Tan, Mai and Thomas, Andrew M and Vincent, Joshua and Mohan, Sreyas and Marcos Morales, Adria and Kulkarni, Shreyas A and Matteson, David S and others},
	journal = Science,
	volume = {387},
	number = {6737},
	pages = {949--954},
	year = {2025}
}

@inproceedings{ZeroIG,
	title={Zero-ig: zero-shot illumination-guided joint denoising and adaptive enhancement for low-light images},
	author={Shi, Yiqi and Liu, Duo and Zhang, Liguo and Tian, Ye and Xia, Xuezhi and Fu, Xiaojing},
	booktitle = CVPR,
	pages = {3015--3024},
	year = {2024}
}

@inproceedings{chen1,
	title        = {High-Fidelity Document Stain Removal via A Large-Scale Real-World Dataset and A Memory-Augmented Transformer},
	author       = {Mingxian Li and Hao Sun and Yingtie Lei and Xiaofeng Zhang and Yihang Dong and Yilin Zhou and Zimeng Li and Xuhang Chen},
	year         = {2024},
	booktitle    = {IEEE Winter Conf. Appl. Comput. Vis.},
	pages        = {7614--7624}
}

@article{chen2,
	author={Guo, Xiaojiao and Chen, Xuhang and Wang, Shuqiang and Pun, Chi-Man},
	journal = CSVT,
	title={Underwater Image Restoration Through a Prior Guided Hybrid Sense Approach and Extensive Benchmark Analysis},
	year = {2025},
	volume = {35},
	number = {5},
	pages = {4784--4800}
}

@article{chen3,
	title={LensNet: An End-to-End Learning Framework for Empirical Point Spread Function Modeling and Lensless Imaging Reconstruction},
	author={Bai, Jiesong and Yin, Yuhao and Dong, Yihang and Zhang, Xiaofeng and Pun, Chi-Man and Chen, Xuhang},
	journal = arXiv,
	note={arXiv:2505.01755},
	year={2025}
}

@article{Interfaces,
	title={Deep Learning Enhanced in Situ Atomic Imaging of Ion Migration at Crystalline--Amorphous Interfaces},
	author={Dong, Weikang and Wang, Yi-Chi and Yang, Chen and Sun, Chunhao and Li, Hesong and Hua, Ze and Wu, Ziqi and Chang, Xiaoxue and Bao, Lixia and Qu, Shuangquan and others},
	journal = NanoLett,
	volume = {24},
	number = {45},
	pages = {14445--14452},
	year = {2024},
}

@inproceedings{pl1,
	title={Boosting image de-raining via central-surrounding synergistic convolution},
	author={Peng, Long and Wang, Yang and Di, Xin and Fu, Xueyang and Cao, Yang and Zha, Zheng-Jun and others},
	booktitle = AAAI,
	volume = {39},
	number = {6},
	pages = {6470--6478},
	year = {2025}
}

@article{pl2,
	title={Lightweight adaptive feature de-drifting for compressed image classification},
	author={Peng, Long and Cao, Yang and Sun, Yuejin and Wang, Yang},
	journal = TMM,
	volume = {26},
	pages = {6424--6436},
	year = {2024},
	publisher={IEEE}
}

@inproceedings{pl3,
	title={Qmambabsr: Burst image super-resolution with query state space model},
	author={Di, Xin and Peng, Long and Xia, Peizhe and Li, Wenbo and Pei, Renjing and Cao, Yang and Wang, Yang and Zha, Zheng-Jun},
	booktitle = CVPR,
	pages = {23080--23090},
	year = {2025}
}

@article{pl4,
	title={PMQ-VE: Progressive Multi-Frame Quantization for Video Enhancement},
	author={Feng, ZhanFeng and Peng, Long and Di, Xin and Guo, Yong and Li, Wenbo and Zhang, Yulun and Pei, Renjing and Wang, Yang and Cao, Yang and Zha, Zheng-Jun},
	journal = arXiv,
	note={arXiv:2505.12266},
	year={2025}
}

@ARTICLE{xj1,
	author={Xiao, Jie and Fu, Xueyang and Zhu, Yurui and Zha, Zheng-Jun},
	journal = PAMI,
	title={Bayesian Window Transformer for Image Restoration}, 
	year={2026},
	volume={48},
	number={3},
	pages={2431--2444},}

@ARTICLE{xj2,
	author={Xiao, Jie and Fu, Xueyang and Liu, Aiping and Wu, Feng and Zha, Zheng-Jun},
	journal = PAMI,
	title={Image De-Raining Transformer}, 
	year={2023},
	volume={45},
	number={11},
	pages={12978--12995},
}

@article{he3,
	title={UnfoldLDM: Deep Unfolding-based Blind Image Restoration with Latent Diffusion Priors},
	author={He, Chunming and Zhang, Rihan and Chen, Zheng and Yang, Bowen and Fang, Chengyu and Lin, Yunlong and Zhang, Yulun and Xiao, Fengyang and Farsiu, Sina},
	journal = arXiv,
	note={arXiv:2511.18152},
	year={2025}
}

@article{DeepRFT,
	title={Deep Residual {Fourier} Transformation for Single Image Deblurring},
	author={Xintian Mao and Yiming Liu and Wei-lei Shen and Qingli Li and Yan Wang},
	journal=arXiv,
	year={2021},
	note={arXiv:2111.11745},
}

@article{zyk1,
	title={Unaligned RGB Guided Hyperspectral Image Super-Resolution with Spatial-Spectral Concordance},
	author={Zhang, Yingkai and Lai, Zeqiang and Zhang, Tao and Fu, Ying and Zhou, Chenghu},
	journal=IJCV,
	volume={133},
	number={9},
	pages={6590--6610},
	year={2025},
	publisher={Springer}
}

@inproceedings{zyk2,
	title={Real Noise Decoupling for Hyperspectral Image Denoising},
	author={Zhang, Yingkai and Zhang, Tao and Nie, Jing and Fu, Ying},
	booktitle=AAAI,
	volume={40},
	number={15},
	pages={12925--12933},
	year={2026}
}

@article{zyk3,
	title={Supervise-assisted Multi-modality Fusion Diffusion Model for PET Restoration}, 
	author={Zhang, Yingkai and Chen, Shuang and Tian, Ye and Gao, Yunyi and Jiang, Jianyong and Fu, Ying},
	journal=TsinghuaSciTechnol,
	year={2026},
	publisher={TUP},
	doi={10.26599/TST.2026.90100027}
}

@article{zyk4,
	title={Enhancing Unregistered Hyperspectral Image Super-Resolution via Unmixing-based Abundance Fusion Learning},
	author={Zhang, Yingkai and Zhang, Tao and Nie, Jing and Fu, Ying},
	journal=arXiv,
	year={2026},
	note={arXiv:2603.07918},
}

@ARTICLE{jyq,
	author={Jiang, Yuqi and Fu, Ying and Liu, Qiankun and Zhang, Jun},
	journal=PAMI, 
	title={MSFA Image Denoising Using Physics-Based Noise Model and Noise-Decoupled Network}, 
	year={2026},
	volume={48},
	number={1},
	pages={859--875}
}

@ARTICLE{lyp1,
	author={Liang, Yingping and Fu, Ying and Hu, Yutao and Shao, Wenqi and Liu, Jiaming and Zhang, Debing},
	journal=PAMI, 
	title={Flow-Anything: Learning Real-World Optical Flow Estimation From Large-Scale Single-View Images}, 
	year={2025},
	volume={47},
	number={10},
	pages={8435--8452}}

@article{lyp2,
	title = {Relation-Guided Adversarial Learning for Data-Free Knowledge Transfer},
	author = {Liang, Yingping and Fu, Ying},
	journal = IJCV,
	year = {2025},
	volume = {133},
	number = {5},
	pages = {2868--2885},
	publisher = {Springer},
}

@ARTICLE{lmy,
	author={Li, Miaoyu and Fu, Ying and Zhang, Tao and Liu, Ji and Dou, Dejing and Yan, Chenggang and Zhang, Yulun},
	journal=PAMI, 
	title={Latent Diffusion Enhanced Rectangle Transformer for Hyperspectral Image Restoration}, 
	year={2025},
	volume={47},
	number={1},
	pages={549--564}}

@ARTICLE{zyh,
		author={Zou, Yunhao and Fu, Ying and Zhang, Yulun and Zhang, Tao and Yan, Chenggang and Timofte, Radu},
		journal=PAMI, 
		title={Calibration-Free Raw Image Denoising via Fine-Grained Noise Estimation}, 
		year={2025},
		volume={47},
		number={7},
		pages={5368--5384}}

@article{CJE5,
		title={A Survey on Degraded Image Segmentation},
		author={Chen, Linwei and Fu, Ying and Shangguan, Jingyu and Xu, Jinglin and Peng, Yuxin},
		journal=CJE,
		volume={1},
		number={1},
		pages={1--20},
		year={2025}
	}

@article{TriMSOD,
		title={TriM-SOD: A Multi-Modal, Multi-Task, and Multi-Scale Spacecraft Optical Dataset},
		author={Zhu, Tianyu and Li, Hesong and Fu, Ying},
		journal=SpaceSciTechnol,
		year={2025},
		volume = {5},
		pages = {0299}
	}

@InProceedings{3D-B2U,
		author="Wang, Jianan and Li, Hesong and Wang, Xiaoyong and Fu, Ying",
		title="3D-B2U: Self-supervised Fluorescent Image Sequences Denoising",
		booktitle=ArtifIntell,
		year="2024",
		pages="130--142",
	}

@article{lhs-small,
		title={Deep-Learning Aided Atomic-Scale Observation of Anisotropic Melting of the Charge Density Wave in TaS2},
		author={Hua, Ze and Qu, Shuangquan and Yan, Li and Dong, Weikang and Zhou, Yaping and Li, Hesong and Chang, Xiaoxue and Bao, Lixia and Wang, Yuhe and Fu, Ying and others},
		journal={Small},
		volume={21},
		number={45},
		pages={e07496},
		year={2025},
		publisher={Wiley Online Library}
	}

@inproceedings{CoordConv,
		author={Liu, Rosanne and Lehman, Joel and Molino, Piero and Such, Felipe Petroski and Frank, Eric and Sergeev, Alex and Yosinski, Jason},
		title={An intriguing failing of convolutional neural networks and the {CoordConv} solution},
		booktitle=NIPS,
		volume={31},
		year={2018}
	}

@inproceedings{CoordAttention,
		author={Hou, Qibin and Zhou, Daquan and Feng, Jiashi},
		title={Coordinate attention for efficient mobile network design},
		booktitle=CVPR,
		pages={13713--13722},
		year={2021}
	}

	

\begin{IEEEbiography}[{\includegraphics[width=1in,height=1.25in,clip,keepaspectratio]{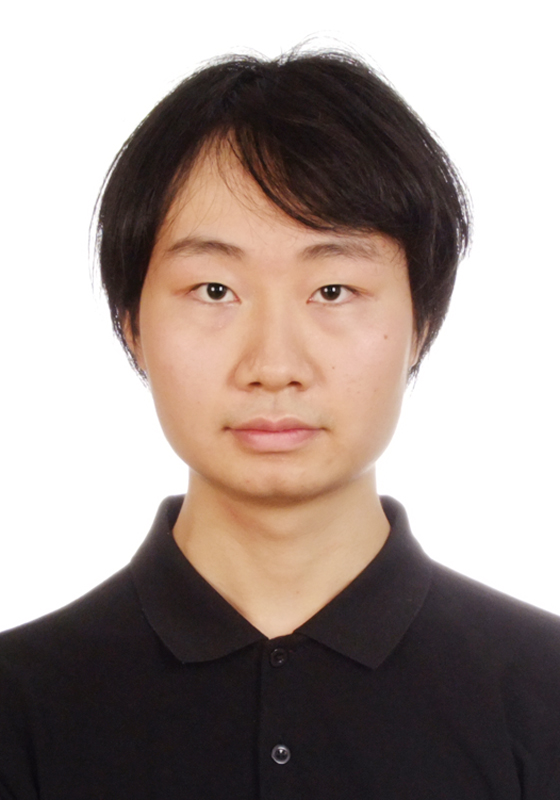}}]{Hesong Li}
	received a B.S. degree in mathematics and applied mathematics from Dalian Maritime University, Dalian, China, in 2021. He is currently a Ph.D. candidate at the School of Computer Science and Technology, Beijing Institute of Technology, Beijing, China. His research interests include artificial intelligence and image processing.
\end{IEEEbiography}

\begin{IEEEbiography}[{\includegraphics[width=1in,height=1.25in,clip,keepaspectratio]{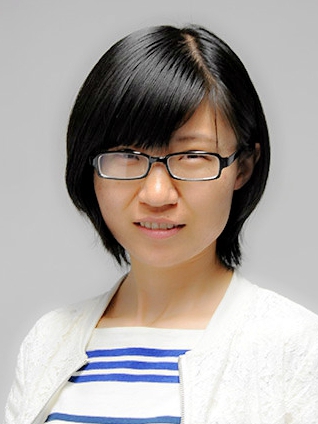}}]{Ying Fu}
	received a B.S. degree in electronic engineering from Xidian University, Xi'an, China, in 2009, an M.S. degree in automation from Tsinghua University, Beijing, China, in 2012, and a Ph.D. degree in information science and technology from the University of Tokyo, Japan, in 2015. She is currently a Professor at the School of Computer Science and Technology, Beijing Institute of Technology. Her research interests include computer vision, image and video processing, and computational photography. 
\end{IEEEbiography}

\begin{IEEEbiography}[{\includegraphics[width=1in,height=1.25in,clip,keepaspectratio]{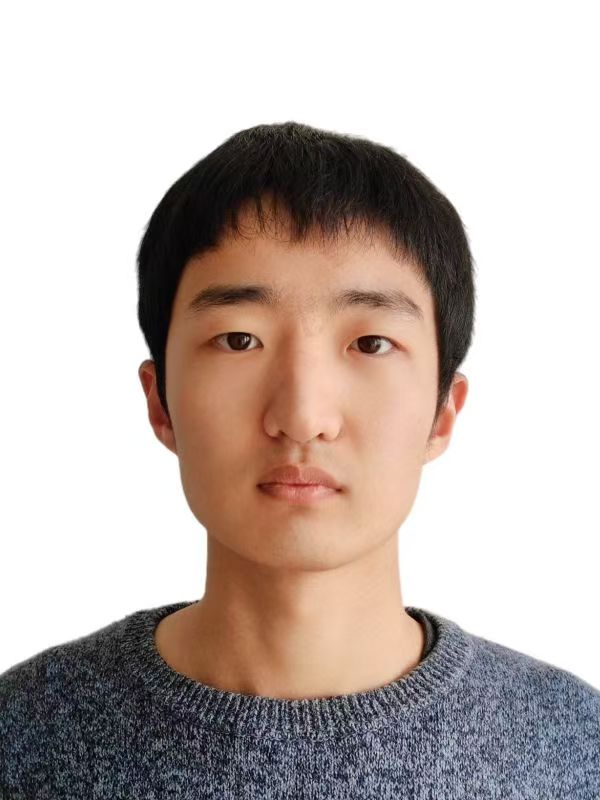}}]{Ziqi Wu}
	received a B.E. degree in automation from Chang'an University, Xi'an, China, in 2022, and an M.E. degree in mechanical engineering from Beijing Institute of Technology, Beijing, China, in 2025. He is currently a Ph.D. candidate at the School of Medical Technology, Beijing Institute of Technology, Beijing, China. His research interests include oxide cathode materials and intelligent electron microscopy-based precise characterization techniques.
\end{IEEEbiography}

\begin{IEEEbiography}[{\includegraphics[width=1in,height=1.25in,clip,keepaspectratio]{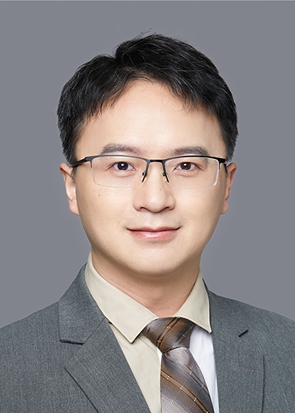}}]{Ruiwen Shao}
	received a B.E. degree in mechanical engineering from Tianjin University, Tianjin, China, in 2010, and a Ph.D. degree in engineering from Beijing University of Technology, Beijing, China, in 2016. He worked as a Postdoctoral Fellow at the Electron Microscopy Laboratory, Peking University, from 2016 to 2018, and as an Assistant Researcher at the Technical Institute of Physics and Chemistry, Chinese Academy of Sciences, from 2018 to 2019. He is currently an Associate Professor at the School of Medical Technology, Beijing Institute of Technology. His research interests include micro-nano robots, micro-nano manipulation, and electron microscopy.
\end{IEEEbiography}

\begin{IEEEbiography}[{\includegraphics[width=1in,height=1.25in,clip,keepaspectratio]{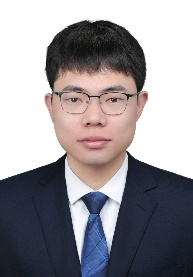}}]{Tao Zhang}
	received B.S. and Ph.D. degrees from the School of Computer Science and Technology, Beijing Institute of Technology, China, in 2017 and 2023, respectively. He is currently an associate professor at Hangzhou Dianzi University. His research interests include intelligent information processing, computational photography, and image processing.
\end{IEEEbiography}

\end{document}